%% file: main.tex
\documentclass[conference]{IEEEtran}
\usepackage[switch]{lineno}
\usepackage{booktabs}
\usepackage{graphicx}
\usepackage{float}  
\usepackage{xcolor}
\usepackage{amsmath} 
\usepackage{listings}
\usepackage{fancyvrb}
\usepackage[numbers]{natbib}
\usepackage{booktabs}
\usepackage{tabularx}
\usepackage{longtable}
\usepackage{array}
\usepackage{algorithm}
\usepackage{algpseudocode}
\usepackage{caption}
\usepackage{subcaption}
\usepackage{url}
\usepackage{doi}
\usepackage{hyperref}
\usepackage{xspace}
\usepackage[most]{tcolorbox}
\newcommand{\method}{\textsc{Prove-RT}\xspace}
\newcommand{\prosa}{\textsc{Prosa}\xspace}
\newcommand{\rocq}{\textsc{Rocq}\xspace}

\newtcolorbox{rqanswerbox}{
  colback=gray!10,
  colframe=gray!60,
  boxrule=0.5pt,
  arc=2pt,
  left=6pt,
  right=6pt,
  top=6pt,
  bottom=6pt
}

\newtcolorbox{compactrqanswerbox}{
  colback=gray!8,
  colframe=gray!45,
  boxrule=0.35pt,
  arc=1pt,
  left=3pt,
  right=3pt,
  top=3pt,
  bottom=3pt,
  before skip=3pt,
  after skip=3pt,
  fontupper=\footnotesize
}

\title{\textsc{Prove-RT}: Generating Mechanized Theorem Prover Scripts for Real-Time Systems using LLMs }
\author{
\IEEEauthorblockN{
Sadat Shahriyar\IEEEauthorrefmark{1},
Shareef Ahmed\IEEEauthorrefmark{2},
Abdullah Al Arafat\IEEEauthorrefmark{1}
}
\IEEEauthorblockA{\IEEEauthorrefmark{1}Florida International University, \IEEEauthorrefmark{2}University of South Florida\\
Email: \{sshahriy, aarafat\}@fiu.edu, shareefahmed@usf.edu}
}

\begin{document}

\maketitle
\input{sections/abstract}
\input{sections/introduction}

\input{sections/challenges}
\input{sections/preliminary}
\input{sections/problemStatement}
\input{sections/methodology}
\input{sections/illustrative_example}
\input{sections/experimental_setup}

\input{sections/evaluation}
\input{sections/discussion}

{\small
\bibliographystyle{IEEEtranN}
\bibliography{references}
}
\clearpage
\onecolumn
\appendix

\input{sections/appendix}

\end{document}

%% file: sections/abstract.tex
\begin{abstract}
Schedulability analysis is essential for certifying real-time systems, but existing tests are often developed through pen-and-paper proofs that are difficult to scale, validate, and maintain. Mechanized verification in \prosa/\rocq offers a rigorous alternative, yet manually constructing such proofs requires substantial domain expertise and proof-engineering effort. Recent successes of large language models (LLMs) across a wide range of tasks make them promising candidates for generating \prosa/\rocq scripts for mechanized theorem provers. However, state-of-the-art LLMs often lack the \prosa-specific knowledge required to correctly use its modeling abstractions and proof patterns.

This paper introduces \method, an LLM-assisted framework for generating \prosa/\rocq scripts to mechanize schedulability analyses in real-time systems literature. \method guides generation through dependency-aware informal sketches, retrieval from processed \prosa documentation, staged skeleton generation, and proof completion. We construct a mechanization-oriented corpus from $1,191$ real-time systems papers, containing $13,134$ informal sketches with dependency information. On a curated evaluation set, direct prompting of state-of-the-art LLMs fails to reliably generate valid \prosa mechanizations, whereas \method achieves a success rate of $44.7\%$. These results show that retrieval-guided and staged LLM assistance can improve automated mechanization of schedulability analysis in \prosa/\rocq.

\end{abstract}

%% file: sections/introduction.tex
\section{Introduction}

Safety-critical real-time systems (RTS) must undergo an offline certification process (typically in the form of a schedulability analysis) to ensure timing correctness at runtime.
Traditionally, schedulability analyses have been devised through pen-and-paper proofs. Although this approach has enabled a rich body of real-time scheduling theory, it is increasingly difficult to sustain as systems become more complex. 

Moreover, manual proofs require careful validation, as subtle mistakes in intermediate lemmas or bounds can affect the final schedulability result and subsequent analyses that build upon it (e.g., \cite{devillers2000liu,bril2006message,nelissen2015timing, chen2019many} are a few papers among many others that identify and address issues in earlier analyses).

An alternative approach is to verify correctness through mechanized proofs. \citet{prosa} developed the first mechanized theorem prover, \prosa, for real-time systems. \prosa provides a \rocq~\cite{rocq2025manual}-based foundation for mechanized schedulability analysis, offering a more rigorous approach for verifying the correctness of pen-and-paper proofs. \prosa is a repository of definitions and proofs for machine-checkable real-time scheduling theory, including task models, schedules, workload functions, interference notions, and schedulability results. By building on \rocq, \prosa enables schedulability analyses to be stated and checked with precision, thereby increasing confidence in the correctness of real-time systems theory.

Unfortunately, mechanized verification does not eliminate the human effort required to construct formal proofs. This is evident from previous works. For example, CertiCAN, a \prosa-based \rocq tool for certifying CAN schedulability-analysis results, required 18,852 lines of \rocq code, excluding the \prosa proofs that it reused~\cite{fradet2023certican}. Similarly, integrating \prosa's verified schedulability analysis into the RT-CertiKOS verified operating-system kernel required 4,135 lines of \rocq for the connection alone, including 1,900 lines for the translation interface to \prosa~\cite{guo2019integrating}. Beyond initial development costs, mechanized proofs also incur maintenance costs: for instance, \citet{bozhko2020abstract} observed that even conceptually simple changes to the underlying model can invalidate existing mechanized proofs and require dozens of person-hours of proof maintenance. These examples suggest that although \prosa provides a reusable foundation for machine-checkable schedulability analysis, manual development and maintenance  remain a substantial barrier to widespread adoption.

To reduce the burden of script preparation for mechanized theorem provers~\cite{rocq2025manual, nipkow2002isabelle, demoura2015lean, norell2007agda} in general cases, a growing body of work has explored automated theorem proving and proof automation using machine learning~\cite{pmlr-v235-blaauwbroek24a,10.1145/3510003.3510138,10.1007/978-3-030-53518-6_17,10.1145/3428299,10.1145/3394450.3397466,10.1145/3593374,gpass}. However, to the best of our knowledge, there has been no prior work on the automated mechanization of schedulability analyses for real-time systems. In this work, we introduce \method, a novel \emph{large-language-model} (LLM)-assisted framework for generating mechanized \prosa proofs for schedulability analyses whose correctness can be verified mechanically.
% The schedulability tests are usually presented as a sequence of definitions, assumptions, lemmas, theorems, and corollaries. 
% \method aims to translate real-time schedulability tests into \prosa scripts with the help of LLM to mechanically verify correctness.    
\emph{It is noteworthy that, beyond mechanizing existing schedulability analyses, \method may also facilitate the trustworthy use of generative AI for developing new schedulability analyses. As generative AI has shown promise across many scientific disciplines~\cite{bfsprover, deepseekproverv2, realprover}, \method can serve as a verification layer for validating AI-assisted or AI-generated schedulability results through mechanized checking.}

\medskip

\noindent\textbf{Challenges and Contributions.} 
Although schedulability analyses are mathematically rigorous, they are typically expressed in a non-mechanized form that lacks the explicit structure required by theorem provers. Therefore, generating \prosa scripts for schedulability analyses differs fundamentally from conventional mathematical proof generation which is more widely studied in the literature.
This introduces three key challenges: (\textit{i}) limited mechanized RTS corpora, restricting LLM understanding of \prosa abstractions and proof patterns; (\textit{ii}) a formalization gap between structurally non-mechanized schedulability analyses and \prosa's explicit proof structure; and (\textit{iii}) the complexity in the structure of schedulability lemmas/theorems compared to mathematical theorem-proving tasks. These challenges are elaborated in Section ~\ref{sec:challenges}.
To overcome them,  \method incrementally mechanizes schedulability analyses through staged formalization, dependency-aware proof construction, and retrieval-augmented grounding using RTS knowledge and \prosa documentation.

In summary, this paper makes the following contributions:
\begin{itemize}
\item We introduce \method, a framework for assisting the mechanization of schedulability analyses in \prosa/\rocq. To the best of our knowledge, \method is the first LLM-assisted framework targeting the \prosa-based mechanization of RTS analysis.

\item We develop a benchmark from $1,191$ real-time systems papers, comprising $13,134$ mechanization-oriented informal sketches and corresponding \prosa/\rocq script artifacts. The benchmark is intended to support future work on LLM-assisted theorem proving and formalization for real-time systems.

\item We analyze the key challenges, recurring failure modes, and corner cases encountered when generating \prosa/\rocq scripts with LLMs. This study provides practical insights for improving future automated mechanization tools for real-time systems analysis.
\end{itemize}

\noindent \textbf{Paper Organization.} The remainder of the paper is organized as follows. Section ~\ref{sec:challenges}  elaborates on the core challenges while mechanizing schedulability analysis with \prosa. Section~\ref{sec:preliminaries} provides background on \prosa and reviews related work. Section~\ref{sec:problem_statement} introduces the notation used throughout the paper and formalizes the problem statement. Section~\ref{sec:method} presents the design of \method. Section~\ref{sec:experimental_setup} describes the construction of the system invariant dataset, the evaluation baselines and metrics, and the implementation details. Section~\ref{sec:motivating_example} provides an example on how the framework works. Section~\ref{sec:eval} presents the evaluation of \method. Finally, Section~\ref{sec:conclusion} concludes the paper by discussing the limitations of \method and outlining directions for future work.

%% file: sections/challenges.tex
\section{Challenges in Mechanizing Schedulability Analysis}
\label{sec:challenges}
Automated theorem proving for real-time schedulability analysis introduces challenges that differ from the more commonly studied setting of LLM-based theorem proving for pure mathematics. In particular, the difficulty is not only to generate a proof script for a given theorem, but also to recover the formal structure needed to express schedulability analyses in \prosa/\rocq.

A key limitation is the scarcity of mechanized training data for real-time schedulability analysis. Recent LLM-based theorem-proving systems benefit from large formal corpora ~\cite{mathlib, minif2f, proofnet, coqgym, holist, putnambench}, which provide many examples of theorem statements, proof scripts, and reusable mathematical libraries. In contrast, public data for mechanized schedulability analysis is very limited. The \prosa library is the main available resource, but its scale is much smaller than mature mathematical proof libraries. This data scarcity limits the ability of LLMs to learn \prosa-specific abstractions, scheduling terminology, type-class assumptions, and proof patterns.

Another challenge is the formalization gap between schedulability analyses as written in the real-time systems literature and the explicit proof structure required by \prosa. Schedulability analyses are usually presented through mathematical notation, prose explanations, assumptions, definitions, intermediate bounds, and proof sketches. Although these presentations are rigorous for human readers, they are not directly mechanizable. A \prosa/\rocq development must explicitly declare variables, hypotheses, type-class instances, section contexts, dependencies, and proof obligations. Therefore, an automated system must first identify the mechanization-relevant constructs, recover their dependencies, normalize notation, and map the extracted concepts to existing \prosa abstractions before proof generation can even begin.

Schedulability-analysis lemmas are also structurally complex. They are often built on a hierarchy of task models, job parameters, arrival constraints, scheduling policies, workload definitions, interference bounds, and response-time properties. These definitions may depend on one another across multiple levels and are usually accompanied by many hypotheses. As a result, the complete formal context needed to state and prove a schedulability lemma in \prosa can span hundreds of lines. This creates a substantial burden on the LLM's context handling and reasoning capabilities.

Beyond the complexity of individual lemmas, \prosa is designed around reusable abstractions and lemmas that apply across different task models and scheduling policies. Whether an existing lemma can be used to prove a target result depends on whether its preconditions hold in the current proof context. Thus, proof generation requires more than selecting tactics: it requires constructing the right context, preserving dependency order, and ensuring that all required assumptions are available. These factors make automated mechanization of schedulability analysis substantially more difficult than direct proof generation for an already formalized theorem statement.

%% file: sections/preliminary.tex
\section{Preliminaries and Related Work}\label{sec:preliminaries}
In this section, we will discuss the necessary \prosa background and the related works on mechanized theorem provers and script generation.

\subsection{Background on \prosa}

\prosa~\cite{prosa} is a \rocq library for mechanized schedulability analysis of RTS. It formalizes real-time scheduling concepts as Gallina (i.e., \rocq specification language) definitions and establishes schedulability results as machine-checked lemmas and theorems. Since \prosa is built in \rocq, a \prosa script follows the standard \rocq development model: required modules are imported, assumptions are introduced through sections and contexts, definitions are stated, and proof obligations are discharged using tactics.

A key feature of \prosa developments is their reliance on explicit proof context. A \texttt{Section} groups together variables, hypotheses, and type-class constraints that are shared by the definitions and lemmas inside it. \prosa uses \rocq type classes to represent reusable modeling assumptions, such as task parameters, job parameters, scheduling policies, and system properties. Although these assumptions may be inferred automatically by \rocq via type-class resolution, the required instances must be available in the current context or imported environment for the script to type-check.

\rocq follows a forward-referencing discipline: every definition, hypothesis, lemma, theorem, and imported module must be introduced before it is used. Thus, a valid \prosa development must be ordered so that all prerequisites of a proof are already available when the proof is checked. This makes dependency ordering and context construction central to writing correct \prosa scripts.

We briefly summarize the \rocq/\prosa notions that are needed to follow the rest of the paper.
% We briefly summarize the \rocq/\prosa notions that are needed to understand the proposed generation pipeline.
% We briefly summarize the \rocq/\prosa notions that are needed to understand the proposed generation pipeline.

\begin{itemize}
    \item \textbf{Proof script} is the \rocq source code used to express a formal development. In \prosa, a proof script contains imports, definitions, assumptions, lemmas, theorems, and tactic-based proofs for schedulability analysis.

    \item \textbf{Proof environment} refers to the collection of formal objects available during proof development. This includes imported libraries, previously defined concepts, declared assumptions, and already-proved lemmas or theorems.

    \item \textbf{Proof context} refers to the local information available at a particular point in a proof. In \prosa, this often includes variables, hypotheses, task and job parameters, scheduling assumptions, and type-class instances introduced inside a section.

    \item \textbf{Proof goal} is the proposition that remains to be proved. A schedulability theorem in \prosa may generate one or more proof goals involving task assumptions, workload bounds, interference bounds, or response-time guarantees.

    \item \textbf{Proof tactic} is a command used to advance a proof by transforming the current goal into simpler subgoals or by solving it directly. Common \rocq tactics include \texttt{intros}, \texttt{apply}, \texttt{rewrite}, \texttt{simpl}, and \texttt{lia}.

    \item \textbf{Proof step} is one application of a proof tactic. Each proof step changes the current proof state and moves the proof closer to completion.

    \item \textbf{Proof state} is the complete status of an interactive proof at a given point, including the current goals and the local context. During proof construction, the proof state changes after each tactic is applied.

    \item \textbf{Type-class resolution} is \rocq's mechanism for automatically finding required instances of abstract interfaces. \prosa uses type classes to represent reusable modeling assumptions, such as job costs, task parameters, arrival information, and scheduling properties.

    \item \textbf{Proof obligation} is a statement that must be proved before a development is complete. Lemmas, theorems, and corollaries introduce proof obligations, whereas definitions mainly introduce formal objects.

    \item \textbf{Compilation} is the process of checking a \rocq/\prosa file. A file compiles only if all referenced objects are available, all statements are well typed, all type-class requirements are resolved, and all proof obligations are completed or explicitly admitted. A \emph{compilation error} occurs when any of these conditions is violated, for example, due to missing imports, undefined variables, unresolved type-class instances, type mismatches, or incomplete proofs. Such errors indicate that the script is not yet a valid mechanized development.
\end{itemize}

\subsection{Related Works}
Mechanized verification provides a rigorous way to validate mathematical and software artifacts by encoding definitions, assumptions, lemmas, and theorems in an interactive theorem prover (ITP), where each proof is checked by a small trusted kernel. Several ITPs, such as \rocq~\cite{rocq2025manual} (commonly known as Coq), Isabelle~\cite{nipkow2002isabelle}, Agda~\cite{norell2007agda}, and Lean~\cite{demoura2015lean}, have been widely used in the verification community. These proof assistants have supported the verification of a broad range of software systems. For example, \rocq has been used to verify the CompCert C compiler~\cite{leroy2009formal}, a lightweight relational database management system~\cite{malecha2010toward}, and distributed systems through the Verdi framework~\cite{wilcox2015verdi}. Isabelle/HOL has been used to verify the seL4 operating-system kernel~\cite{klein2009sel4}. More recently, Lean has been used to formalize and verify neural networks through TorchLean~\cite{george2026torchlean}.

In real-time systems, \prosa provides a \rocq-based foundation for mechanized schedulability analysis ~\cite{prosa}. It formalizes core real-time scheduling concepts including task models, schedules, workload functions, interference notions, and schedulability results. Building on this foundation, prior work has used \prosa to mechanize and validate several schedulability-analysis results including response-time analysis, CAN schedulability certification, FIFO scheduling, busy-window reasoning, and connections between response-time analysis and network calculus ~\cite{bozhko2020abstract, fradet2023certican, bedarkar2022intuition, maida2022foundational, roux2022formal}. These works demonstrate the value of mechanized verification for increasing confidence in real-time systems theory. However, they still require substantial manual proof engineering and domain expertise.

To reduce the burden of manual proof development, prior approaches have proposed various methods for generating proof tactics, selecting relevant premises, and guiding proof search using learned models~\cite{pmlr-v235-blaauwbroek24a,10.1145/3510003.3510138,10.1007/978-3-030-53518-6_17,10.1145/3428299,10.1145/3394450.3397466,10.1145/3593374,gpass}. More recently, large language models (LLMs) have shown promise in generating formal proofs and assisting interactive theorem proving~\cite{han2022proof,NEURIPS2022_377c2531,NEURIPS2023_44414694,jiang2021lisa,Polu2020GenerativeLM,rango}. However, these efforts have largely focused on general-purpose theorem proving or domains such as mathematics and software verification. To the best of our knowledge, there has been no prior effort to automatically generate \prosa proofs for mechanized schedulability analysis from real-time systems papers. 

%% file: sections/problemStatement.tex
\section{Framework Model and Problem Statement}
\label{sec:problem_statement}

\subsection{Framework Model}
We first define the main objects used throughout the \method pipeline. 

\noindent\textbf{Schedulability Analysis ($A$).} Let $A$ be a schedulability analysis for an RTS scheduling problem. Such an analysis is typically presented using mathematical definitions, assumptions, lemmas, theorems, corollaries, and proof arguments. Unlike standard \rocq tactic-generation tasks, $A$ is not given as a proof goal inside an existing proof environment. Instead, it is a schedulability test whose formal constructs, dependency structure, and proof obligations must be recovered before mechanization.

\noindent\textbf{System Invariant ($I$). }We use the term \emph{system invariant} to refer to any mechanization-relevant construct in $A$ that must be represented in the final \prosa/\rocq development. A system invariant may be a definition, hypothesis, assumption, lemma, theorem, corollary, fix-point, or other formal claim required to establish the correctness of the schedulability test. We denote the set of extracted system invariants by
\(
I = \{i_1, i_2, \ldots, i_n\}.
\)

\noindent\textbf{Informal Sketch ($\{K_j\}$). }An informal sketch $K_j$ describes the statement of the invariant $i_j$, the intuition behind it, and the intended proof outline. These sketches serve as the intermediate representation between the natural-language schedulability analysis and the generated \prosa/\rocq script. Each invariant $i_j \in I$ is associated with an informal sketch $K_j$. 

\noindent\textbf{Dependency Structure ($G_D$). }The extracted invariants form a dependency structure. We represent this structure as a directed acyclic graph (DAG)
\(G_D = (I, D),\)
where each node corresponds to a system invariant and each directed edge represents a dependency, e.g., an edge \((i_p, i_j) \in D\) means that invariant $i_j$ depends on $i_p$. Thus, $i_p$ must appear earlier than $i_j$ in the generated \prosa/\rocq script. The DAG, therefore, defines the order in which invariants should be formalized. This ordering is required because \rocq is forward-referencing: all definitions, assumptions, lemmas, theorems, and imports must be introduced before they are used.

\noindent\textbf{\prosa Documentation Corpus ($P$). }Let $P$ be the processed \prosa documentation corpus. $P$ contains documentation fragments, definitions, assumptions, reusable lemmas, and example proof structures extracted from the \prosa library. For each invariant $i_j$, the retrieval module selects a relevant subset
\(R_j \subseteq P\)
to provide the library-specific context needed for script generation.

The generated script is built incrementally according to the dependency order induced by $G_D$. 

\subsection{Problem Statement}

Given a schedulability analysis $A$ and a processed \prosa documentation corpus $P$, the goal of \method is to generate a dependency-ordered \prosa/\rocq script $S$ that mechanizes the schedulability analysis and compiles under the \prosa/\rocq environment as:
% . Using the notation introduced in the framework model, the overall transformation is written as
\[
M : (A, P) \rightarrow S,
\]
where $M$ denotes the \method generation pipeline and
\(S = \{s_1, s_2, \ldots, s_n\}\)
is the generated \prosa/\rocq script. Each code block $s_j \in S$ is intended to formalize the corresponding system invariant $i_j \in I$.

A successful output must satisfy three requirements. First, each generated code block $s_j$ must correctly represent the corresponding extracted invariant $i_j$. Second, the ordering of code blocks in $S$ must respect the dependency graph $G_D=(I,D)$, so that every prerequisite invariant is introduced before any invariant that depends on it. Third, the final script $S$ must compile under the \prosa/\rocq environment.

We decompose this objective into two subproblems. 

\noindent\textbf{Skeleton Generation (via $M_{\mathrm{skel}}$). }The first subproblem is \emph{skeleton generation}. For each invariant $i_j$, the skeleton-generation module uses the informal sketch $K_j$, the retrieved \prosa context $R_j$, the dependency graph $G_D$, and the partial \prosa/\rocq script already generated for earlier invariants, denoted by $S_{<j}$, to produce a structurally valid \prosa/\rocq code block:
\[
M_{\mathrm{skel}} : (K_j, R_j, G_D, S_{<j}) \rightarrow s_j.
\]

After the skeleton-generation stage, some generated \prosa/\rocq fragments contain proof-bearing constructs, such as lemmas, theorems, or corollaries, whose proof bodies were intentionally left as \texttt{Admitted.} We refer to these unfinished proof bodies as deferred proof obligations. The proof-completion stage attempts to replace each \texttt{Admitted.} placeholder with a valid \rocq proof.

For each deferred proof associated with $s_j$, the proof-completion module takes the current partial script $S_{<j}$, the corresponding informal sketch $K_j$, and the retrieved \prosa context as input $R_j$, and generates a proof fragment $\pi_j$ intended to replace the \texttt{Admitted.} placeholder.

\noindent\textbf{Proof Completion (via $M_{\mathrm{proof}}$). }The second subproblem is \emph{proof completion}. For each deferred proof in $s_j$, the proof-completion module uses the current partial script $S_{<j}$, the informal sketch $K_j$, and the retrieved \prosa context $R_j$ to generate a proof fragment $\pi_j$:
\[
M_{\mathrm{proof}} : (S_{<j}, s_j, K_j, R_j) \rightarrow \pi_j.
\]

The generated proof fragment $\pi_j$ is intended to replace the corresponding \texttt{Admitted} placeholder. It is accepted only if the resulting \prosa/\rocq script is checked successfully by the \rocq compiler. If compilation fails, compiler feedback or proof-state information is used to guide repair.

Thus, the problem addressed by \method is broader than next-tactic prediction. It requires transforming a real-time systems schedulability analysis into a dependency-ordered \prosa/\rocq script that can be mechanically checked by the proof assistant.

%% file: sections/methodology.tex
\section{Prove-RT Framework}
\label{sec:method}
\begin{figure}[t]
    \centering
    \includegraphics[width=0.9\linewidth]{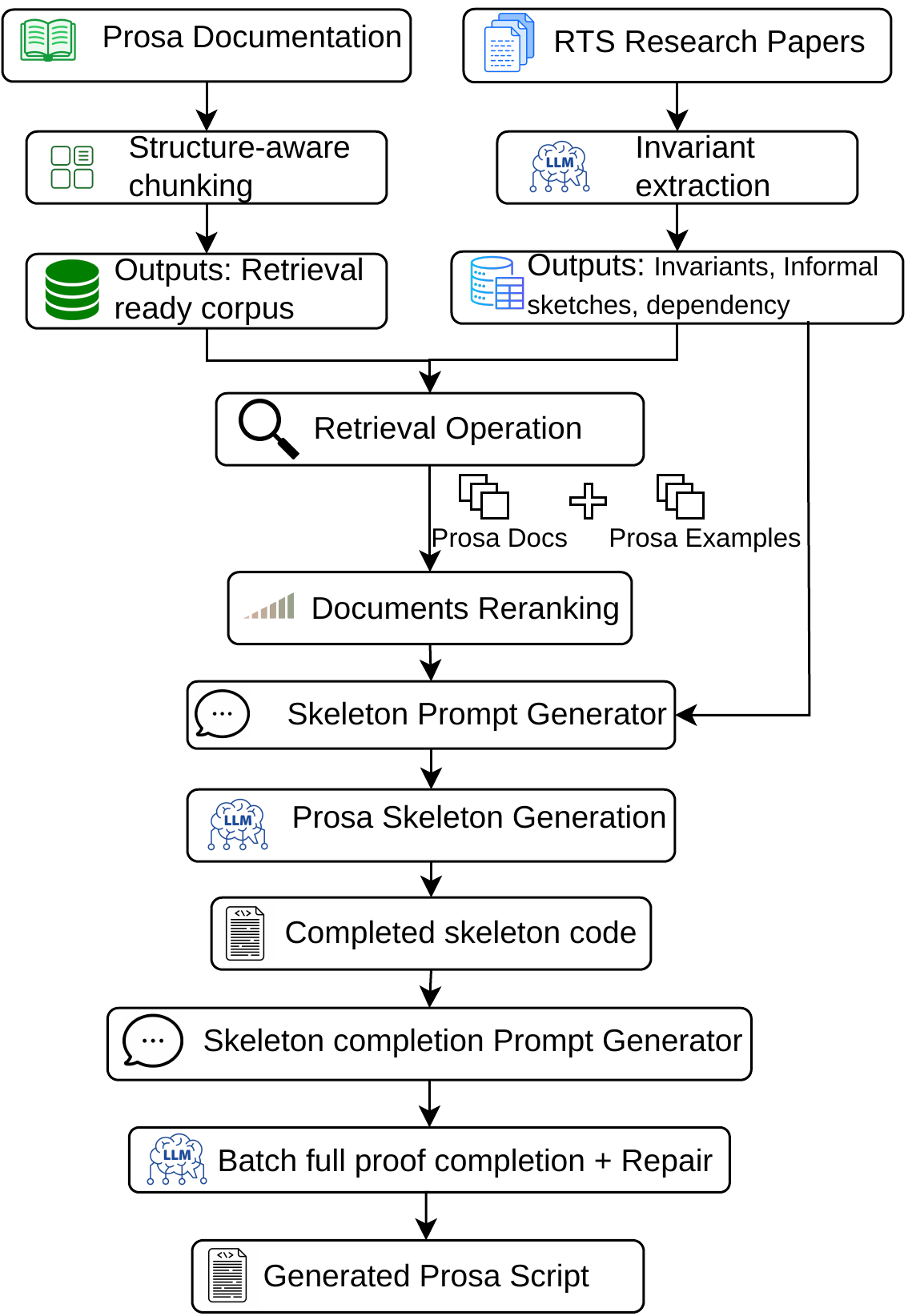}
    \caption{Overview of the proposed \prosa/\rocq proof-generation pipeline.}
    \vspace{-3mm}
    \label{fig:system_diagram}
\end{figure}

\method consists of five stages: (1) Extracting system invariants, informal sketches, and dependency graph from formally written schedulability analysis; (2) Processing the \prosa documentation into retrieval-ready fragments; (3) retrieving relevant documentation and examples, with dependency recovery for proof-oriented modules; (4) generating and validating a structurally correct proof skeleton with deferred obligations; and (5) completing the deferred proofs using iterative repair. We explain these stages in the following subsections. Fig.~\ref{fig:system_diagram} illustrates an overview of the \method. 

\subsection{System Invariant Extraction}
\noindent\textbf{Extraction of Invariants and Sketches. }
The first stage of \method transforms the input, schedulability analysis $A$, into the intermediate representation required for mechanization, consisting of a set of system invariants $I$, their corresponding informal sketches $\{K_j\}$, and a dependency graph $G_D=(I,D)$. This stage is necessary as schedulability analyses in the RTS literature are usually written for mathematical presentation rather than direct mechanization. As a result, the constructs needed for \prosa/\rocq generation may appear in different forms, including prose explanations, equations, definitions, lemmas, theorems, claims, or corollaries.

To facilitate this process, \method first prompts an LLM to identify candidate system invariants from schedulability test $A$.
The LLM is guided using carefully designed prompts that include practical examples and detailed guardrails to ensure consistency and accuracy in the output.

For each extracted invariant $i_j \in I$, the LLM generates an informal sketch $K_j$. Informal sketches provide a structured, step-by-step description of each system invariant in plain text, capturing both the logical flow and the intended proof outline. An example informal sketch is outlined in Listing~\ref{lst:informal_sketch} in Appendix ~\ref{app:illustrative_example_informal_sketch}.

\medskip

\noindent\textbf{Construction of Dependency Graph.}
The extracted dependency information is used to construct the dependency graph $G_D=(I,D)$. For each invariant $i_j$, the LLM identifies other invariants that must be established before $i_j$ can be formalized. Each such relation is represented as an edge in $D$. The extracted invariants are then ordered according to this graph so that, when the \prosa/\rocq script is generated, prerequisites are introduced before the constructs that depend on them. The resulting unit-level informal sketches, ordered by $G_D$, form the primary input to the subsequent stages of \method.

\subsection{\prosa Documentation Processing}
\label{sec:prosa_doc_processing}
The second stage of \method constructs the processed \prosa documentation corpus $P$ used for retrieval. This stage is necessary because direct LLM-based generation of \prosa/\rocq scripts is difficult without library-specific context. The \prosa library relies on domain-specific abstractions, type-class assumptions, reusable lemmas, and proof conventions that are often not available to a pretrained model. Thus, the \prosa documentation is processed into retrieval-ready fragments so that later stages can retrieve relevant proof constructs for each invariant $i_j$.

The \prosa documentation is organized into six main modules:
\textit{Analysis}, \textit{Behavior}, \textit{Implementation}, \textit{Model},
\textit{Results}, and \textit{Util}. These modules are used to guide the construction of $P$. In particular, the \textit{Results} module is
used primarily as a source of complete proof examples, since it contains
formally verified schedulability results. The remaining modules provide
supporting context, including system definitions, modeling assumptions, formal
definitions, reusable lemmas, and auxiliary proof components.

As the modules differ in both structure and purpose, they are processed using
two chunking strategies. 

\medskip

\noindent\textbf{Proof-Oriented Chunking.} For the \textit{Analysis} and \textit{Results} modules,
proof-oriented chunks are used which pairs each formal code block with its
corresponding descriptive text. This preserves the connection between a proof
component and the explanation that motivates it, allowing retrieval to operate
over meaningful proof-level units such as definitions, lemmas, and theorems. During retrieval, the retrieved proof-level units are augmented with their relevant dependencies, so that lemmas, theorems, and definitions are provided together with the supporting constructs required for script generation. This dependency reconstruction step is described in the next section~\ref{sec:retrieval_process}.

\medskip

\noindent \textbf{Section-Level Chunking.} For the remaining modules, section-level chunks are used, since these modules are
typically organized around smaller and more self-contained concepts, assumptions,
or helper components.

The resulting set of documentation fragments forms the processed corpus $P$. During retrieval, \method selects a relevant subset $R_j \subseteq P$ for each invariant $i_j$, which provides the LLM with \prosa-specific context for generating the corresponding \prosa/\rocq code block $s_j$.

\subsection{Retrieval Process}
\label{sec:retrieval_process}

\noindent\textbf{Query Construction.} The retrieval stage constructs the context $R_j \subseteq P$ for each invariant $i_j$. Given the informal sketch $K_j$, the goal is to retrieve the most relevant fragments from the processed \prosa documentation corpus $P$ so that the skeleton-generation module has access to library-specific definitions, assumptions, lemmas, and example proof structures.

Each informal sketch may contain information about multiple invariants, including a target invariant and the prerequisites on which it depends. To preserve the dependency order induced by $G_D$, we process the sketches one invariant at a time. For each invariant $i_j$, its corresponding sketch $K_j$ is used as the retrieval unit.

For each informal sketch unit $K_j$, three components are used as retrieval queries: the statement of the invariant, the intuition underlying it, and the conclusion it establishes. Together, these components capture both the formal objective and the supporting reasoning of the target proof step.

Retrieval is performed independently for the statement, intuition, and conclusion queries, producing a ranked top-$k$ list for each component. The retrieved candidates are then merged, and the highest-scoring documentation sections are selected as the retrieval context $R_j$.

\medskip

\noindent\textbf{Dependency Recovery for Proof-Oriented Modules.} For candidates retrieved from the \textit{Analysis} or \textit{Results} modules,  an additional syntax-aware dependency recovery step is applied before adding them to $R_j$. These modules contain proof-oriented files in which later sections often rely on earlier definitions, hypotheses, lemmas, or typeclass contexts. Since \rocq follows a forward-referencing discipline, a section with index $N$ can depend only on preceding sections with indices from $0$ to $N-1$ within the same file. \method exploits this ordering to recover the earlier sections needed to make the retrieved proof fragment usable in downstream generation.

The dependency recovery step assigns different weights to occurrences of identifiers based on their syntactic context. Identifiers before the \texttt{Proof.} keyword are treated as the strongest signals because they appear in the type-level statement of the declaration. Explicit references following tactics such as \texttt{apply} and \texttt{rewrite} are treated as medium-strength signals, while other proof-level tokens receive lower weight because they are more likely to include tactic noise or external-library names. If an identifier appears in multiple contexts, its maximum weight is kept to preserve the strongest dependency signal without double counting.

After extracting candidate identifiers from the three zones, each identifier is assigned a weight
\[
w(x)=\max \bigl( \alpha \cdot \mathbf{F}[x\in Z_A],\; \beta \cdot \mathbf{F}[x\in Z_B],\; \gamma \cdot \mathbf{F}[x\in Z_C] \bigr),
\]
Here, $Z_A$, $Z_B$, and $Z_C$ denote the sets of identifiers extracted from the type-level zone, the explicit-reference zone, and the remaining proof-token zone, respectively. The parameters $\alpha$, $\beta$, and $\gamma$ are weighting coefficients assigned to these zones, with $\alpha > \beta > \gamma$ to reflect the relative strength of the dependency signal provided by each syntactic context. To improve robustness, tokens are filtered out that are unlikely to correspond to meaningful local dependencies, such as grammar keywords, logical connectives, single-character tokens, self-references, and proof-local names introduced during tactics.

After extracting identifiers, they are matched against declarations in earlier sections of the same file. A section receives a score when it defines a matched identifier, weighted by the identifier's dependency score; sections with nonzero scores are treated as likely direct dependencies. Unmatched identifiers contribute a small score to the imports file to prevent external-library references from being ignored. Because direct dependencies may themselves depend on earlier constructs, the procedure is applied recursively and arranges the recovered sections in their original order so that each construct appears before it is used.

The recovered dependencies are assembled with the retrieved section before being included in $R_j$. Imports are placed at the top level, code fragments are ordered for compilation, and unicode operators are normalized when needed. A lean version is first compiled containing only the recovered dependencies to reduce context size and noise. If this fails, a full-context version is used that includes all predecessor sections from the same file, which helps capture implicit dependencies such as typeclass requirements.

For the remaining modules, the documentation is more naturally organized into self-contained sections. Therefore, each section is treated as a single retrieval unit, the top-$k$ matching sections are retrieved, and used directly without additional dependency recovery.

\subsection{Skeleton Code Generation}

The skeleton-generation stage addresses the first subproblem defined in Section~\ref{sec:problem_statement}. For each invariant $i_j$, it generates a structurally valid \prosa/\rocq code block $s_j$ using the invariant's informal sketch $K_j$, the retrieved context $R_j$, the dependency graph $G_D$, and the partial script $S_{<j}$ generated for earlier invariants.

This stage deliberately separates structure generation from proof generation. Generating the correct type signature of a \prosa/\rocq construct, including its name, variable bindings, typeclass constraints, and statement, is generally more tractable for a language model than simultaneously discovering a valid proof strategy.  Therefore, the skeleton stage focuses on producing well-typed declarations and statements, while proof bodies are deferred to the proof-completion stage. This decomposition provides an intermediate artifact that can be checked by \texttt{coqc}, allowing \method to detect structural errors before attempting proof synthesis.

\medskip

\noindent\textbf{Prompt Construction.} For each invariant $i_j$, we construct the skeleton-generation prompt from four inputs: the informal sketch $K_j$, the retrieved \prosa context $R_j$, the dependency graph $G_D$, and the partial script $S_{<j}$. These components provide the model with the relevant library context, examples of \prosa structure, and the current proof context. The informal sketch $K_j$ for the target section is appended last so that it remains the immediate generation objective.

In addition to this contextual information, the prompt includes two generation rules to ensure that the generated block $s_j$  is structurally meaningful. First, \textsc{proof-skeleton rule} is enforced where every proof-bearing construct must be generated with a deferred proof obligation using \texttt{Admitted.}, and no proof tactics are allowed at this stage. Thus, the model may generate either of the following forms:
\begin{center}
\begin{minipage}{0.5\linewidth}
\footnotesize
\begin{verbatim}
Lemma foo : <statement>.
Proof.
  Admitted.
\end{verbatim}
\end{minipage}
\end{center}
or
\begin{center}
\begin{minipage}{0.5\linewidth}
\footnotesize
\begin{verbatim}
Lemma foo : <statement>.
Admitted.
\end{verbatim}
\end{minipage}
\end{center}
Non-proof-bearing constructs, such as definitions, declarations, and type-level specifications, must instead be generated in full because they do not create deferred proof obligations and are required for the skeleton to type-check.

Second, \textsc{assumption-integrity rules} are enforced. In preliminary experiments, a common failure mode was that the model made the proof artificially easy by turning claims that should be proven into unsupported assumptions. To prevent this, the prompt explicitly prohibits introducing unsupported hypotheses, restating the target claim as an assumption, or encoding the proof obligation in a vacuous way. Claims from $K_j$ must remain proof-bearing constructs, while genuine preconditions must be encoded as part of the corresponding formal statement.

These rules make skeleton checking more reliable. If the generated block $s_j$ fails to compile, the error is more likely to indicate a structural problem, such as an incorrect type, undefined identifier, or missing import, rather than a failed or misleading proof attempt. They also prevent the model from producing speculative proof bodies that may appear syntactically valid but are semantically incorrect and difficult to repair later.

\begin{figure}[t]
    \centering
    \includegraphics[width=\columnwidth]{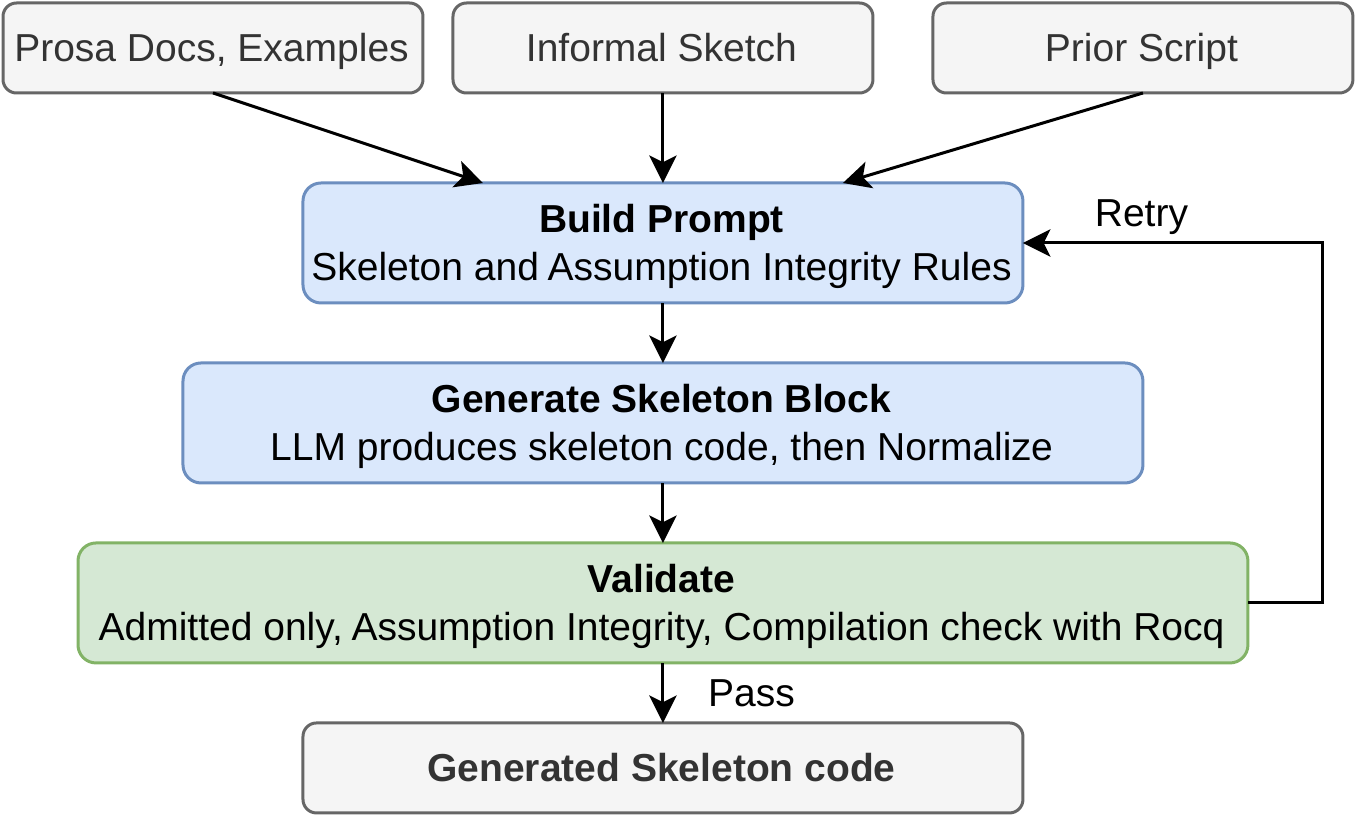}
    \caption{Skeleton code generation}
    \vspace{-3mm}
    \label{fig:Skeleton_code_generation}
\end{figure}

\medskip

\noindent\textbf{Validation and Retry.} After each generation attempt, the produced skeleton $s_j$ is validated in two steps. First, it is checked that all proof-bearing constructs follow the admitted-only rule and contain no completed or partial proof bodies. Second, $s_j$ is appended to the current partial script $S_{<j}$ and the resulting script is compiled with \texttt{coqc}. This verifies structural validity and type correctness.

Non-proof-bearing constructs are accepted directly because they do not open a \texttt{Proof.} block. If validation or compilation fails, the section is regenerated; sections that repeatedly fail are logged and skipped so that the pipeline can continue.

\subsection{Skeleton Code Completion}
\label{sec:skeleton-completion}

After skeleton generation, each code block $s_j$ may contain proof-bearing constructs whose proof bodies are deferred using \texttt{Admitted.} The goal of the completion stage is to generate a proof fragment $\pi_j$ that replaces each deferred proof and makes the resulting \prosa/\rocq script compile. For each deferred proof, the completion module uses the current partial script $S_{<j}$, the corresponding informal sketch $K_j$, and the retrieved \prosa context $R_j$ as input. In this way, the system builds directly on the previous stage: the skeleton provides the formal statement and context, while the completion stage focuses only on filling in the missing proofs. 

\medskip

\noindent\textbf{Batch Completion with Iterative Repair.} \method prompts LLM to generate the full proof body for a deferred proof in one response. The generated proof fragment $\pi_j$ replaces the corresponding \texttt{Admitted.} placeholder in $s_j$, and the resulting partial script is then checked with the \rocq compiler. Furthermore, to ensure that the generated script proves the intended target lemma rather than circumventing it, we include a proof-integrity checker. The checker validates the LLM-generated proof and detects whether the model has modified the original problem, introduced unsupported auxiliary facts, or shifted the main proof obligation into a separate construct. Edits outside the intended proof region are treated as invalid. Concretely, the checker flags constructs such as \texttt{Axiom}, \texttt{Parameter}, \texttt{Parameters}, \texttt{Conjecture}, \texttt{Conjectures}, \texttt{Admitted}, \texttt{admit}, and \texttt{Abort}. It also detects whether the LLM introduces new global lemmas or definitions outside the proof region that effectively carry the main proof burden. Such modifications are rejected because they may allow the script to compile without actually proving the original lemma. At the same time, the checker permits harmless changes that do not alter the meaning of the theorem, such as importing additional trusted libraries from \prosa or MathComp etc. If the script compiles successfully and the proof-integrity checker reports no violations, the script is accepted.

If compilation fails or any violation is reported by the proof-integrity checker, \method enters an iterative repair loop in which compiler feedback guides corrections. At each iteration, the error message and its location are extracted from the compiler output and incorporated into a repair prompt together with  the informal sketch $K_j$ and the retrieved context $R_j$. The LLM then proposes a revised proof fragment, which is inserted into the section and checked again.

This process is attractive because it allows the model to generate an entire proof in one step, while still benefiting from compiler-guided repair when the initial attempt is incorrect.

\begin{figure}[t]
    \centering
    \includegraphics[width=0.8\columnwidth]{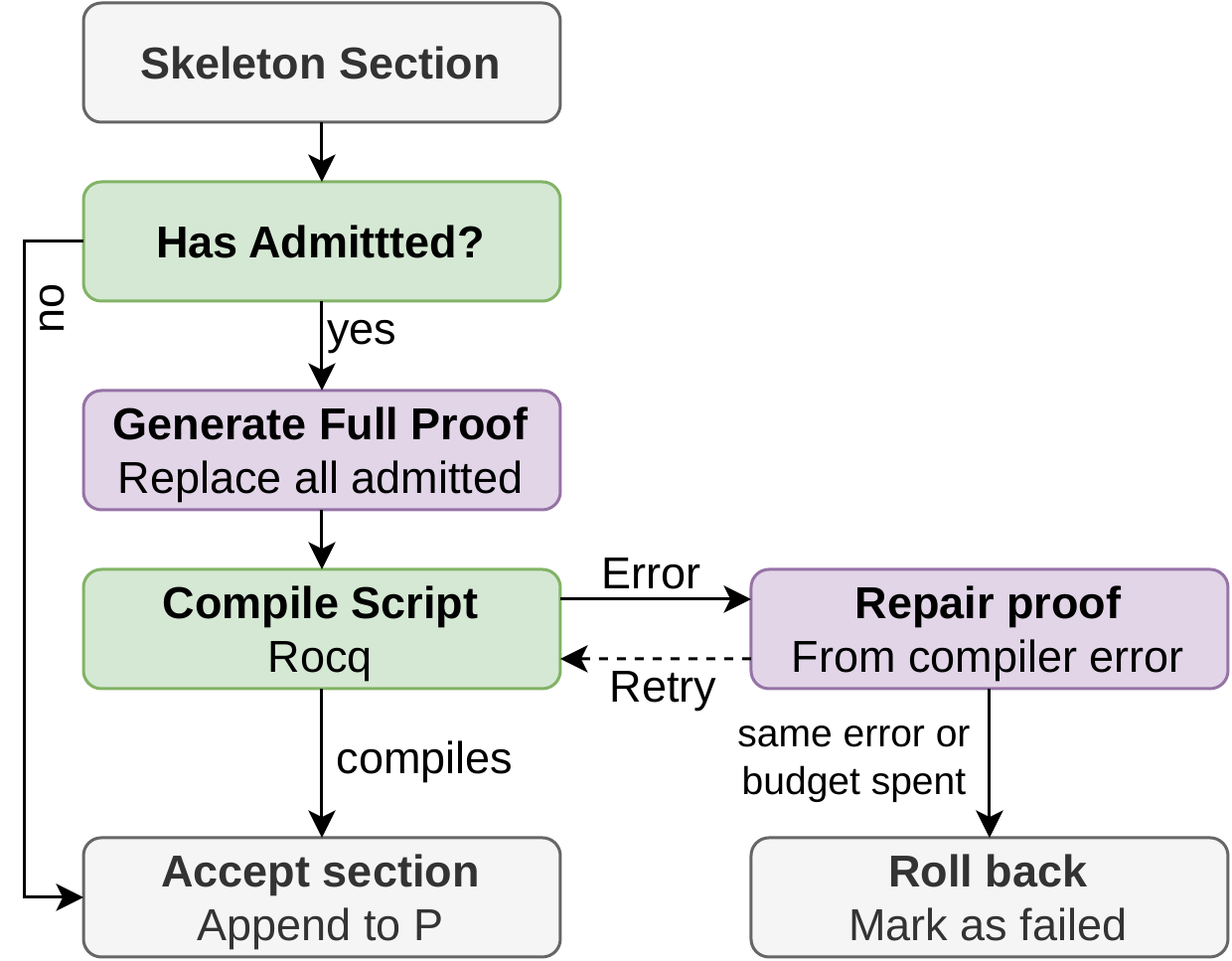}
    \caption{Batch Completion with Iterative Repair}
    \vspace{-4mm}
    \label{fig:Skeleton_code_generation}
\end{figure}

\medskip

%% file: sections/illustrative_example.tex
 
\section{Illustrative Example}
\label{sec:motivating_example}
 
We illustrate the end-to-end workflow of \method using a concrete schedulability analysis drawn from the RTS literature, starting with the extraction of intermediate representations and leading to the final machine-verified \prosa script.

\subsection{Source Material and Extraction}
 
We consider the paper \emph{``Worst-Case Timing Requirements of Real-Time Tasks with Time Redundancy''}~\cite{wctr} as a
representative example. From this paper, we extract the following
definition and claim, which characterize the worst-case timing
requirements of a task under fault-tolerant execution. The formal notations introduced in the paper are as follows:
\noindent \textsc{Worst-Case Timing Requirement:}
Let $a_i(k_i)$ denote the additional processing time and runtime
overhead required to tolerate $k_i$ faults during the mission time of a
task~$\tilde{T}_i$. The worst-case timing requirement $W_i(k_i)$ of
$\tilde{T}_i$ is given by
\begin{equation}\label{eq:wctr_general}
  W_i(k_i) = W_i(0) + A_i(k_i),
\end{equation}
where $A_i(k_i)$ denotes the worst-case value of $a_i(k_i)$ and
$W_i(0)$ represents the failure-free worst-case execution time (WCET).
 
\noindent \textsc{Retry-Based Redundancy:}
Assume that a task restarts from the beginning after each fault, with no
rollback recovery. Let $R^*$ denote the constant restart overhead. Then
the worst-case timing requirement under the retry mechanism is
\begin{equation}\label{eq:wctr_retry}
  W_i(k_i) = k_i \cdot (R^* + W_i(0)) + W_i(0).
\end{equation}

\medskip
\noindent
These results are automatically extracted and structured using
LLM. The extraction process identifies the formal statement,
variables, assumptions, and conclusions associated with each invariant.
In addition, it also produced an \emph{informal sketch}. We provide the complete JSON extraction and the corresponding informal sketch in Appendix~\ref{app:illustrative_example_informal_sketch}.
 
A notable property of the extraction is that it is
\emph{dependency-aware}. For each invariant, we identify the
previously introduced invariants on which it depends; this information is
recorded explicitly in the extracted representation. These dependencies
induce a partial order that governs the subsequent formalization
pipeline: an invariant is formalized only after all of its dependencies
have been processed. In this example, Claim~1 depends on Definition~1,
so the definition must first be formalized. Since \prosa follows
an interpreted execution model in \rocq, all prerequisite definitions and
constructs must be available before they are referenced, making
dependency-aware ordering essential for correct compilation.
 
\subsection{Retrieval-Augmented Generation}
 
Given the informal sketch of an invariant, the next step is to translate
it into a \prosa script. To support this translation, we employ
retrieval-augmented generation (RAG) over the \prosa codebase.
We use the \emph{statement}, \emph{conclusion}, and \emph{intuition}
fields from the informal sketch as query, retrieve their independent results and keep the top-$k$ results from two complementary sources: (i)~\emph{example
scripts} that demonstrate similar constructs or proof patterns, and
(ii)~\emph{documentation fragments} describing the syntax, semantics,
and usage conventions of relevant \prosa modules. Together,
these provide the LLM with sufficient context to generate correct and
idiomatic \prosa script.
 
\subsection{Skeleton Code Generation}
 
Using the retrieved context, we prompt the LLM to generate
\emph{skeleton code}—a structurally complete, type-checkable
\prosa script in which all imports, section boundaries,
type-class contexts, variable declarations, and definition bodies are
fully specified, while every proof obligation is replaced by
\texttt{Admitted.} Listing~\ref{lst:skeleton} shows the skeleton code
generated for the motivating example.
 
\begin{lstlisting}[language=ML, basicstyle=\ttfamily\footnotesize,caption={Skeleton code generated from the informal sketch. All structural elements are fully elaborated; the proof body is deferred via \texttt{Admitted.}}, label={lst:skeleton}]
Require Export prosa.util.all.
Require Export prosa.behavior.time.
Require Export prosa.model.task.concept.
Require Export prosa.model.aggregate.workload.
 
Section WorstCaseTimingRequirement.
 
  Context {Task : TaskType}.
  Context `{TaskCost Task}.
 
  Context {Job : JobType}.
  Context `{JobTask Job Task}.
  Context `{JobCost Job}.
 
  Variable W_i_0 : work.
  Variable k_i : nat.
  Variable R_star : work.
 
  Definition W_i_k_i (A_i_k_i : work) : work :=
    W_i_0 + A_i_k_i.
 
  Lemma W_i_k_i_retry :
    W_i_k_i (k_i * (R_star + W_i_0))
    = k_i * (R_star + W_i_0) + W_i_0.
  Proof.
  Admitted.
 
End WorstCaseTimingRequirement.
\end{lstlisting}
 
The skeleton faithfully encodes the structure of both invariants.
Definition~1 is realized as the function \texttt{W\_i\_k\_i}, which
takes the fault-tolerance overhead \texttt{A\_i\_k\_i} as an argument
and returns its sum with the failure-free WCET \texttt{W\_i\_0}. The
\prosa type \texttt{work}, defined as \texttt{nat}, represents
discrete units of processor service. Claim~1 is stated as
\texttt{Lemma~W\_i\_k\_i\_retry}, which asserts that instantiating
\texttt{W\_i\_k\_i} with the retry-specific overhead yields exactly
Equation~\eqref{eq:wctr_retry}. The type-class contexts
(\texttt{TaskType}, \texttt{JobType}, \texttt{TaskCost}, etc.)\ anchor
the formalization within \prosa's modeling framework and ensure
that the definitions are compatible with the library's broader
infrastructure.
 
The \texttt{Admitted.} directive instructs \rocq to accept the lemma
statement without proof, allowing the entire file to type-check
successfully. This confirms that the formalization structure—imports,
scoping, type-class resolution, and the lemma statement itself—is sound
before any proof synthesis is attempted.
 
\subsection{Proof Completion}
 
In the final phase, we prompt the LLM a second time to discharge the
\texttt{Admitted} obligations. This two-phase decomposition is
deliberate: by supplying the complete skeleton as context, the LLM gains
visibility into the surrounding definitions, type-class instances, and
any auxiliary lemmas, enabling it to generate proof tactics that are
consistent with the broader formalization. The prompt includes the
retrieved context from the RAG step, the full skeleton code, and an
instruction to complete a specific \texttt{Admitted} block. For this
example, the LLM produces the following proof:
 
\begin{lstlisting}[language=ML, basicstyle=\ttfamily\footnotesize, caption={Completed proof of the retry-based worst-case timing requirement}, label={lst:proof}]
  Lemma W_i_k_i_retry :
    W_i_k_i (k_i * (R_star + W_i_0))
    = k_i * (R_star + W_i_0) + W_i_0.
  Proof.
    unfold W_i_k_i.
    lia.
  Qed.
\end{lstlisting}
 
The proof proceeds in two steps. First, \texttt{unfold W\_i\_k\_i}
$\delta$-reduces the definition, exposing the underlying goal:
\[
  \texttt{W\_i\_0} + k_i \times (R^* + \texttt{W\_i\_0})
  = k_i \times (R^* + \texttt{W\_i\_0}) + \texttt{W\_i\_0}.
\]
This is an equality over natural numbers that follows directly from the
commutativity of addition. The \texttt{lia} tactic, which implements a
decision procedure for linear integer arithmetic, discharges it
automatically. The completed script compiles under \rocq and confirms that
the retry-based worst-case timing requirement is a valid instantiation
of the general fault-tolerant WCET model formalized in \prosa.
 
\medskip
\noindent
This example demonstrates how our pipeline systematically transforms
informal real-time scheduling results into machine-verified
\prosa proofs through a structured sequence of extraction,
retrieval-augmented skeleton generation, and targeted proof completion. A second illustrative example can also be found in Appendix~\ref{app:illustrative_example_2}.

%% file: sections/experimental_setup.tex
\section{Experimental Setup}
\label{sec:experimental_setup}
We evaluate \method by investigating the following research questions (RQs).

\noindent\textbf{RQ1:} To what extent can \method formalize schedulability tests as \prosa/\rocq scripts?

\noindent\textbf{RQ2:} How does the dependency depth of informal sketches influence the success of mechanization?

\noindent\textbf{RQ3:} How does the choice of retrieval method affect the effectiveness of \method?

\subsection{System Invariant Dataset Construction}
\label{sec:system_invariant_dataset_construction}

\begin{figure}[t]
    \centering
    \includegraphics[width=\linewidth]{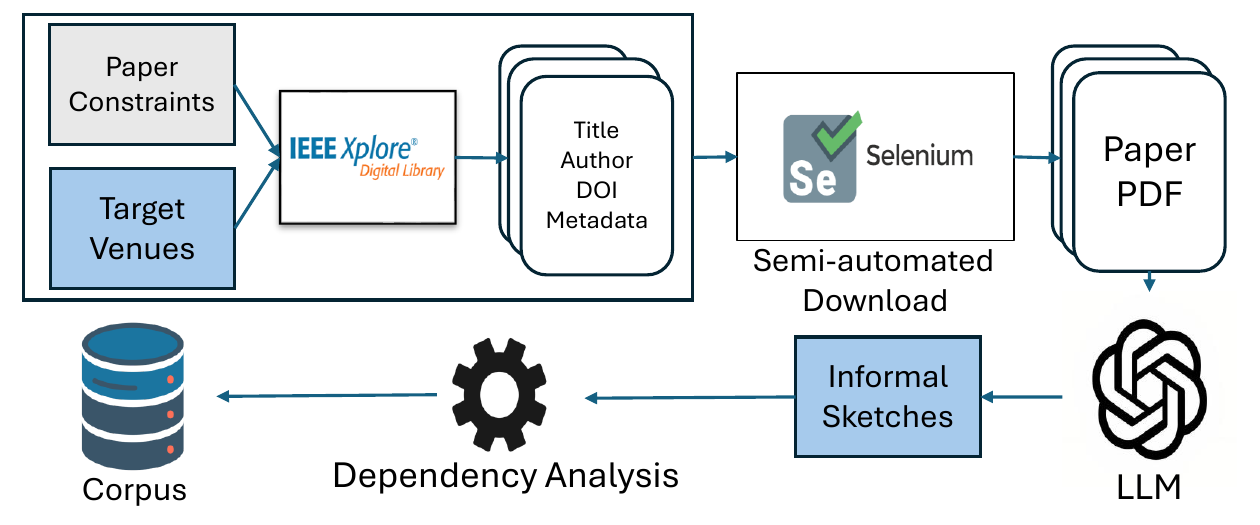}
    \caption{System Invariant Dataset Collection}
    \vspace{-4mm}
    \label{fig:system_invariant_system_diagram}
\end{figure}

As one of the main artifacts of this work, \method System Invariant Dataset is constructed, which is a large-scale collection of schedulability-analysis invariants designed to support and evaluate LLM-assisted \prosa/\rocq mechanization. The dataset provides instances of the framework objects introduced earlier: schedulability analyses $A$, extracted invariant sets $I$, informal sketches $K_j$, and dependency graphs $G_D=(I,D)$. Figure~\ref{fig:system_invariant_system_diagram}   presents a high-level overview of the system invariant corpus collection process.

To construct the source corpus for system-invariant extraction, schedulability-analysis papers were collected from established real-time systems and embedded systems venues. The IEEE Xplore API was used to retrieve paper metadata. The search was restricted to major venues, including RTSS, RTAS, ECRTS, EMSOFT, RTCSA, RTNS, RSS, and IROS. The full search query and venue list are provided in Appendix~\ref{app:ieee_query}.

The API returned 1,991 paper records. For each record, the metadata was extracted to locate the corresponding full-text PDF. Since the IEEE Xplore API does not directly support bulk full-text PDF downloads, a semi-automated browser-assisted workflow was used with personal access credentials. For each paper, the PDF URL was extracted from the retrieved metadata and was opened in an authenticated browser session using \texttt{Selenium}. Then \texttt{Selenium} was used to interact with the Chrome PDF viewer and trigger the download action automatically. Because authenticated sessions may expire during long download runs, the workflow also included a session-recovery mechanism that restored access through automated browser interactions. After retrieval, duplicate files caused by overlapping searches or repeated download sessions were removed, resulting in approximately 1,870 unique papers.

Each collected PDF was then converted into a structured XML representation using \texttt{GROBID}. The XML format preserves document structure, making it more suitable for LLM-based extraction than raw PDF text. This structured representation was used as input to the invariant extraction stage of \method. In our implementation, we used Gemini-2.5-Flash as the LLM and prompted it to identify candidate system invariants $I$ from each schedulability analysis and to extract their summaries, informal sketches $K_j$, and dependency information for constructing $G_D$. The prompts included examples and guardrails to encourage consistent output and to distinguish mechanization-relevant constructs from general explanatory text.

After Gemini extraction, a deterministic validation pass was applied to filter structurally invalid outputs before constructing the final dataset. The validator constructed an inter-invariant dependency graph from the extracted \texttt{identifier} and \texttt{dependencies} fields, where each invariant is represented as a node, and each resolved dependency is represented as a directed edge from the dependent invariant to its prerequisite. To resolve dependency references, the validator uses a three-stage matching process: exact, normalized, and fuzzy matching. Exact matching requires the dependency string to match an extracted identifier character-for-character. Normalized matching canonicalizes both dependency strings and candidate identifiers by lowercasing text, replacing formatting artifacts such as backticks and underscores with spaces, removing non-alphanumeric punctuation, and collapsing repeated whitespace. If both exact and normalized matching fail, fuzzy matching compares the normalized dependency string against all normalized candidate identifiers using a character-level similarity score and links it only to the highest-scoring candidate when the score exceeds $0.88$. This step is intended to recover near-duplicate identifiers with minor residual formatting differences while avoiding spurious dependency edges.

Using the resolved dependency graph, the validator checks for unresolved dependencies, self-dependencies, forward references, and dependency cycles. Papers are labeled as \textit{keep}, \textit{review}, or \textit{reject}: papers with dependency cycles or an unresolved-dependency ratio above $0.15$ are rejected, papers with weaker structural issues are marked for review, and structurally valid papers are retained. After this filtering step, the retained corpus contained $1,191$ papers and $13,134$ informal sketches.

Another important characteristic of the retained system invariant dataset is the distribution of extracted constructs by type. As discussed earlier, each unit-level informal sketch contains a target invariant together with the dependencies required to establish it. Consequently, a single sketch may include multiple system invariants that must be introduced and proved sequentially before the final target invariant can be mechanized. We therefore analyze the full set of system invariants appearing across all collected sketches.

This analysis reveals substantial variation in how authors formulate and name system invariants. Across the collected papers, 73 distinct raw invariant types were identified. To make these invariants compatible with \rocq-based proof development, they were normalized into six broader categories corresponding to supported proof constructs: definitions, hypotheses, lemmas, theorems, corollaries, and fixpoints. The complete fine-grained mapping from raw sketch kinds to \rocq keywords is provided in Appendix~\ref{app:raw_sketch_mapping}. Overall, the extraction process produced 13,134 normalized system invariants, including their dependency information. Table~\ref{tab:invariant_category_distribution} reports the distribution of invariants across the normalized categories.

We further analyze the number of invariants appearing in each unit-level informal sketch. Each sketch contains a target invariant together with the dependencies that must be established before the target can be mechanized. 

The resulting corpus is both an input to \method and a reusable artifact for studying LLM-assisted mechanization of real-time systems schedulability analyses. The distribution is described in Appendix~\ref{app:invariant_dist}.

\subsection{Baselines}
\method is compared against direct LLM-based generation of \prosa/\rocq scripts. Since recent LLMs have shown strong capability in proof synthesis and \rocq code generation, and since \prosa is built on top of \rocq, these baselines evaluate whether LLMs can mechanize real-time systems schedulability analyses without the additional guidance introduced by \textsc{PROVE-RT}.

We consider two baseline settings. First, the LLM is prompted to mechanize a target invariant using only the corresponding text from the original schedulability-analysis paper, without external \prosa documentation or skeleton-code generation. Second, the LLM is provided with the extracted informal sketch for the target invariant, but receives no retrieval support and does not use the skeleton-code generation stage. These baselines measure how far direct LLM generation can go without dependency-aware retrieval, skeleton generation, and proof-repair mechanisms.

\subsection{Evaluation Metrics}

\method counts a generated script as successful only if it is accepted by the \rocq proof checker; that is, the completed proof must compile without errors, contain no remaining deferred proof obligations, and pass the proof-integrity checker without any reported violations. This follows the standard proof-assistant acceptance criterion used in prior \rocq proof-generation work, where generated proofs are considered correct only when they lead the theorem prover to proof termination with \texttt{Qed}~\cite{gpass,rango}.

To measure the effectiveness of \method, we use \textbf{success rate}, a standard metric used in prior work to evaluate \rocq code and proof generation~\cite{coqgym,ddafv,tactoc}.

Success rate is measured as the fraction of proof constructs for which the tool generates a successful proof script:
\begin{equation}
\text{Success Rate} = \frac{N_S}{N_T},
\label{eq:success_rate}
\end{equation}
where $N_S$ denotes the number of successfully mechanized sketches with successful proof scripts, and $N_T$ denotes the total number of evaluated sketches.

\subsection{Implementation Details}
For the system invariant extraction stage, \textsc{Gemini-2.5-Flash} was employed as the backbone large language model. The model is prompted with structured, guardrailed instructions alongside few-shot examples to ensure consistent XML parsing.

For the retrieval-augmented generation (RAG) component, the retrieval corpus was constructed from the CoqDoc-generated HTML documentation of \textsc{Prosa}, yielding $5,097$ documentation fragments from $356$ source files. Three retrieval strategies were evaluated: BM25, dense retrieval, and hybrid retrieval. For dense retrieval, the \texttt{nomic-embed-code-7b} was used to generate vector embeddings and store them in a FAISS vector index. For each query, the top-$K$ most relevant documentation chunks were selected, with $K=5$, and they were provided as context to the LLM during skeleton generation.

The proof completion pipeline uses  \texttt{Claude-Opus-4.6}. It is executed in two phases. In the first phase, the skeleton code is synthesized with \texttt{Admitted.} directives to verify type-checking and type-class resolution. In the second phase, the LLM is prompted to discharge individual proof obligations using automation tactics for linear integer arithmetic. All interaction with the interactive theorem prover is managed via an automated script running \textsc{Rocq} version 9.1.0.

\begin{table}[t]
\centering
\caption{Distribution of extracted elements by category}
\label{tab:invariant_category_distribution}
\small
\setlength{\tabcolsep}{4pt}
\renewcommand{\arraystretch}{1.05}
\begin{tabular}{lrrr}
\toprule
\textbf{Category} & \textbf{Invariant Count} & \textbf{Percentage} & \textbf{\# Raw Kinds} \\
\midrule
Definition & 6817 & 51.9\% & 17 \\
Lemma      & 4144 & 31.6\% & 38 \\
Theorem    & 1822 & 13.9\% & 1  \\
Corollary  & 228  & 1.7\%  & 1  \\
Fixpoint   & 115  & 0.9\%  & 14 \\
Hypothesis & 8    & 0.1\%  & 2  \\
\midrule
\textbf{TOTAL} & \textbf{13134} & \textbf{100.0\%} & \textbf{73} \\
\bottomrule
\end{tabular}
\vspace{-4mm}
\end{table}

The dataset construction and semi-automated PDF collection workflows are executed via Selenium driving an authenticated Chrome browser session. Text extraction and document structuring are processed using \texttt{GROBID} to convert raw PDFs into structured XML. The entire pipeline is implemented in Python 3.10 and evaluated on an Ubuntu 24.04 LTS server equipped with an Intel Xeon w5-3423 processor with 12 cores and 24 hardware threads, and 64GB of system RAM.

%% file: sections/evaluation.tex
\section{Evaluation}
\label{sec:eval}
We evaluate \method according to the three research questions introduced in Section~\ref{sec:experimental_setup}.

\noindent \textbf{Evaluation of RQ1.} To evaluate the effectiveness of \method, we compare it against direct \prosa/\rocq script generation using state-of-the-art LLMs, including GPT and Claude. Although these models have demonstrated strong general capabilities in generating \rocq code, their ability to generate \prosa scripts for mechanizing schedulability analyses remains unclear. We therefore conduct a pilot study to assess how well these models perform in this domain-specific setting. We then evaluate how \method guides these models for producing mechanically checkable \prosa/\rocq scripts.

For this evaluation, we performed a human-in-the-loop curation step to select a representative subset from the retained invariant corpus. The selection focused on scheduling-analysis categories that are well aligned with existing \prosa abstractions while still spanning different levels of mechanization difficulty. These categories include uniprocessor fixed-priority response-time analysis, uniprocessor EDF response-time or demand-bound analysis, FIFO/FCFS response-time analysis, non-preemptive and limited-preemptive analyses, blocking and resource-sharing analyses, self-suspending task analyses, multiprocessor global EDF/global fixed-priority analysis, and multiprocessor partitioned scheduling analysis. After curation, the final evaluation corpus contained $109$ papers and $1,904$ unit-level informal sketches.

From this curated evaluation corpus, we further selected a smaller evaluation subset of $300$ unit-level informal sketches using proportional stratified sampling over dependency-depth categories. Let $n_c$ denote the number of sketches in category $c$, and let $N$ denote the total number of sketches in the curated corpus. For each category, we computed the sampling quota as
\[
q_c = 300 \cdot \frac{n_c}{N}.
\]
We first selected $\lfloor q_c \rfloor$ sketches from each category and then assigned the remaining slots to the categories with the largest fractional remainders. Within each category, sketches were sampled uniformly at random using a fixed seed of $42$ for reproducibility.

\begin{table}[t]
\centering
\caption{Overall mechanization success across generation modes.}
\label{tab:construct_category_success}
\small
\setlength{\tabcolsep}{4pt}
\renewcommand{\arraystretch}{1.1}
\begin{tabular}{lcc}
\toprule
\textbf{Mode} & \textbf{Formalized} & \textbf{Success Rate} \\
\midrule
Paper Statements + GPT-5              & 0/300  & 0.0\% \\
Informal Sketch + GPT-5               & 0/300  & 0.0\% \\
Informal Sketch + Claude-Opus-4.6      & 1/300  & 0.33\% \\
\textsc{PROVE-RT-hybrid}                     & 123/300 & 41.0\% \\
\textsc{PROVE-RT-bm25}                     & 126/300 & 42.0\% \\
\textsc{PROVE-RT-dense}                     & \textbf{134/300}  & \textbf{44.7\%} \\
\bottomrule
\end{tabular}
\vspace{-4mm}
\end{table}
\method achieves the best performance among the evaluated approaches when used with dense RAG, mechanizing $134$ informal sketches and achieving a success rate of $44.7\%$. This result is encouraging because automated formalization remains challenging even in more established proof-assistant settings; prior neural theorem-proving systems such as Rango and GPass report success rates of approximately $32\%$ and $35\%$, respectively~\cite{rango,gpass}. Although these results are not directly comparable due to differences in benchmarks and proof domains, they provide useful context for interpreting the difficulty of the task. Given that \method operates in the specialized and low-resource setting of \prosa-based real-time systems mechanization, a $44.7\%$ success rate indicates substantial progress.

Other retrieval methods also perform competitively within \method: \textsc{BM25} mechanizes $126$ sketches with a success rate of $42.0\%$, while hybrid retrieval mechanizes $123$ sketches with a success rate of $41.0\%$. In contrast, direct generation with GPT fails to mechanize any construct, whether prompted with paper statements or informal sketches. Claude mechanizes only $1$ sketch from the informal sketches, corresponding to a success rate of $0.33\%$. These results suggest that direct prompting, even with structured informal sketches, is insufficient for reliable \prosa/\rocq generation. The detailed comparison is shown in Table~\ref{tab:construct_category_success}.

An interesting observation from the direct-generation baselines is that both GPT and Claude often produce compilable \rocq scripts without actually using \prosa. In these cases, the generated scripts may type-check in \rocq, but they do not rely on the definitions, abstractions, or verified results provided by the \prosa library. Since our goal is to mechanize real-time systems schedulability analyses within \prosa, we count such outputs as failures.

In our study, Claude produced compilable scripts for $39$ of the $300$ informal sketches. However, only $1$ of these scripts used \prosa; the remaining $38$ were generic compilable \rocq scripts and were therefore discarded. The behavior was even more pronounced for GPT, it generated $148$ compilable \rocq scripts out of $300$ attempts, but none of them used \prosa. These results further highlight the limitation of direct LLM prompting, even when the generated output is syntactically valid and type-correct in \rocq, it may fail to mechanize the target schedulability analysis in the intended \prosa framework.

This confirms that decomposing the task into sketch-guided retrieval, skeleton generation, and proof completion improves the reliability of LLM-assisted \prosa mechanization.

\begin{rqanswerbox}
\noindent\textbf{Observation I.}
\method is substantially more effective than direct LLM prompting for \prosa/\rocq mechanization. On the $300$ sampled informal sketches, \textsc{Prove-RT-dense} achieves the best result, mechanizing $134$ sketches with a success rate of $44.7\%$. 
% \textsc{Prove-RT-bm25} and \textsc{Prove-RT-hybrid} also perform competitively, achieving $42.0\%$ and $41.0\%$ success rates, respectively. 
By contrast, direct generation with GPT-5 produces no valid \prosa mechanizations, and Claude-Opus-4.6 succeeds on only $1$ sketch ($0.33\%$). This demonstrates that successful mechanization requires more than general \rocq generation ability; it requires \prosa-aware retrieval, dependency-aware structuring, and staged proof generation.
\end{rqanswerbox}

\begin{figure*}[t]
    \centering
    \begin{subfigure}[t]{0.32\textwidth}
        \centering
        \includegraphics[width=\linewidth]{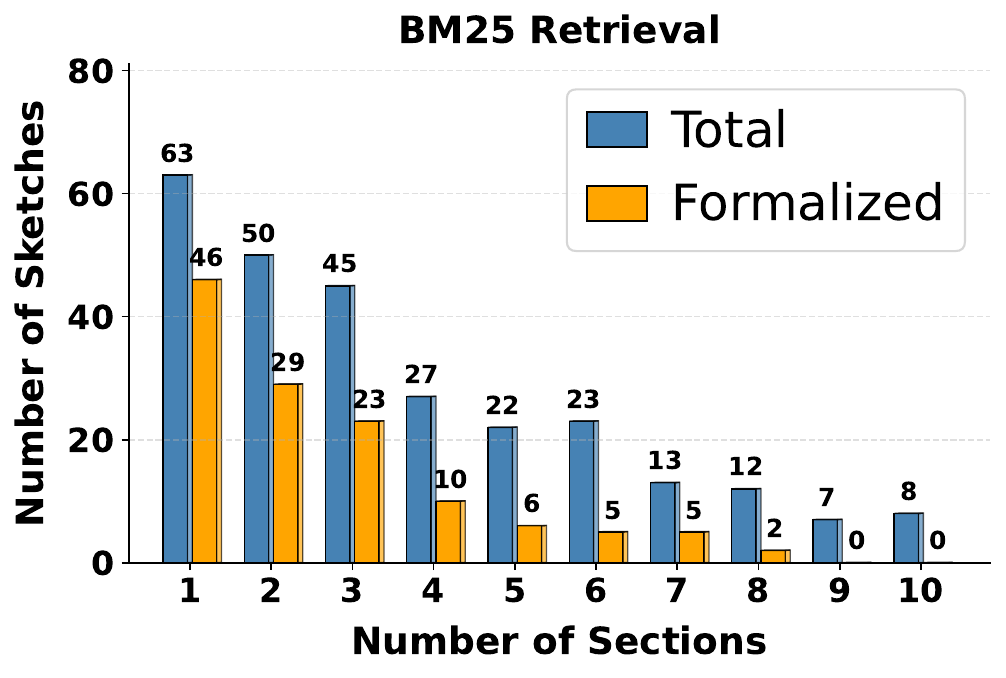}
        \caption{BM25}
        \label{fig:section_success_bm25}
    \end{subfigure}
    \hfill
    \begin{subfigure}[t]{0.32\textwidth}
        \centering
        \includegraphics[width=\linewidth]{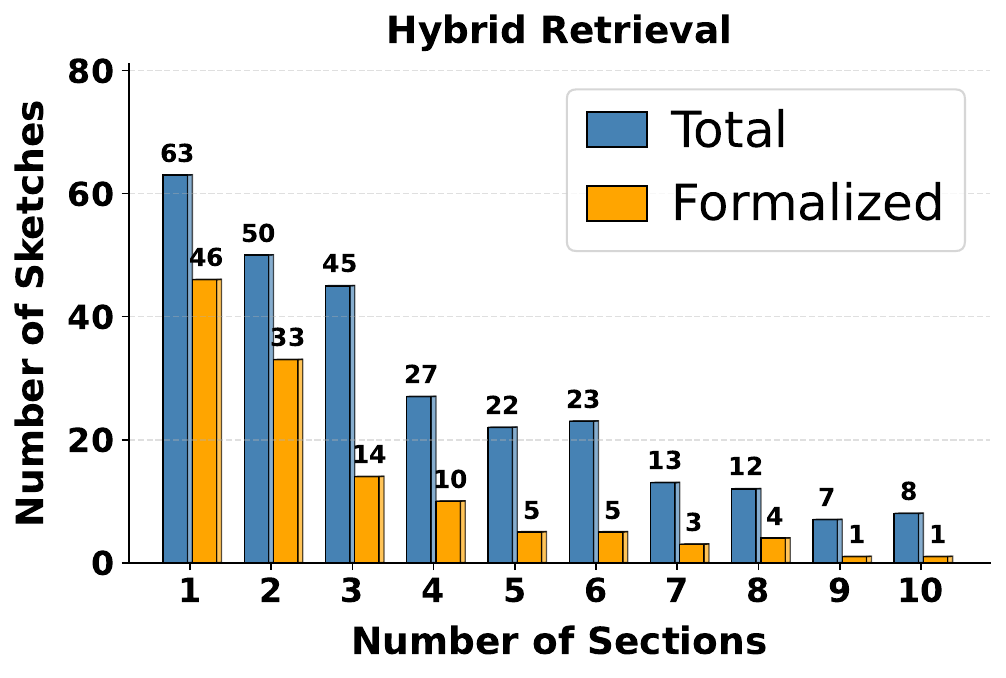}
        \caption{Hybrid}
        \label{fig:section_success_hybrid}
    \end{subfigure}
    \hfill
    \begin{subfigure}[t]{0.32\textwidth}
        \centering
        \includegraphics[width=\linewidth]{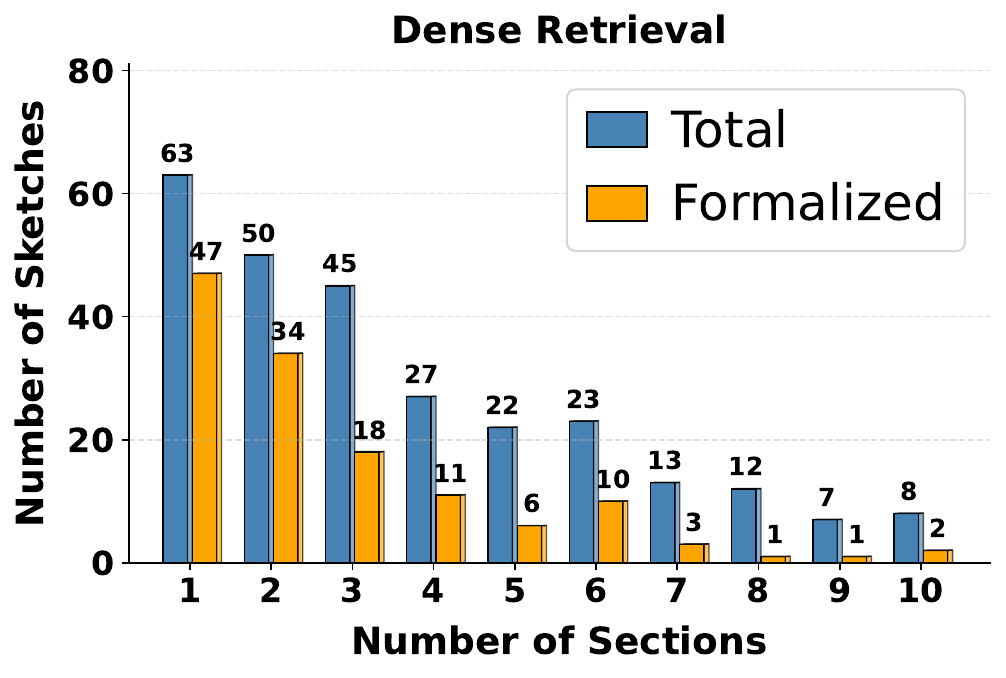}
        \caption{Dense}
        \label{fig:section_success_dense}
    \end{subfigure}

    \caption{Mechanization success by dependency depth for different retrieval methods. For readability, the plot shows sketches with at most $10$ sections. For each dependency depth, the blue bar shows the total number of sketches and the orange bar shows the number successfully formalized by \textsc{Prove-RT}.}
    \vspace{-4mm}\label{fig:section_success_by_rag}
\end{figure*}

\noindent\textbf{Evaluation of RQ2. }To evaluate how the structural complexity of an informal sketch affects mechanization, we group the $300$ sampled informal sketches by the number of sections they contain. In our dataset, each sketch is decomposed into a sequence of sections, where later sections may depend on definitions, assumptions, or intermediate results introduced earlier in the sketch. Thus, the number of sections serves as a proxy for the dependency depth and mechanization complexity of the sketch. For each group, we compare the total number of sketches against the number of sketches that \method successfully formalizes as compilable \prosa/\rocq scripts.

Figure~\ref{fig:section_success_by_rag} shows the distribution of total and successfully formalized sketches across different section counts. Each pair of bars corresponds to sketches with the same number of sections: the first bar shows the total number of sketches in that group, while the second bar shows how many of them were successfully mechanized by \method.

The trend suggests that section count is a useful indicator of mechanization difficulty. Sketches with fewer sections typically correspond to simpler formalization tasks, such as isolated definitions or short proof obligations, where the generated \prosa/\rocq code depends on limited prior context. In contrast, sketches with many sections require \method to preserve a longer chain of definitions, assumptions, and intermediate results. This increases the likelihood that an error in an earlier generated code block propagates to later blocks and causes the final script to fail compilation. Consequently, failures in high-section-count sketches often reflect compound challenges, such as missing or incorrect prerequisites, mismatched identifiers, or insufficient retrieved context. At the same time, successful cases among multi-section sketches show that \method can mechanize nontrivial dependency structures rather than only isolated constructs or single-step proof obligations.

It is important to note that the number of sections is only an approximate measure of complexity. Some short sketches may contain difficult proof obligations, while some longer sketches may consist mostly of definitions or straightforward intermediate claims. Nevertheless, section count provides a useful aggregate view of how dependency depth affects end-to-end mechanization success.

\begin{rqanswerbox}
\noindent\textbf{Observation II.}
The success of \method decreases as sketches contain more sections, suggesting that deeper dependency chains make mechanization harder. Nevertheless, \method succeeds on several multi-section sketches, indicating that its dependency-aware extraction, retrieval, and staged generation strategy can support nontrivial schedulability-analysis mechanization.
\end{rqanswerbox}

\medskip

\noindent\textbf{Evaluation of RQ3.}
Prior work such as ~\cite{rango} uses BM25-style sparse retrieval to retrieve relevant proof context. Sparse retrieval is a natural baseline because it is effective when the query and target documents share exact identifiers, theorem names, library symbols, or domain-specific terminology. However, the retrieval setting in \method is different. The \prosa documentation contains both formal code blocks and descriptive text explaining the logical role of definitions, lemmas, assumptions, and proof patterns. Moreover, for each code block, we generate an additional LLM-based description so that the retrieval corpus captures not only the surface syntax of the code but also its intended semantic role.

This motivates our evaluation of different retrieval methods. In \method, retrieval queries are derived from the informal sketch $K_j$, which describes the statement, intuition, and conclusion of a target invariant in natural language. These queries often do not share exact tokens with the corresponding \prosa documentation or code identifiers, even when they refer to the same concept. Dense retrieval is therefore useful because it maps both sketch-derived queries and documentation descriptions into a shared semantic embedding space. At the same time, sparse retrieval may still be useful for recovering exact \prosa identifiers, theorem names, and library-specific terminology.

\begin{table}[t]
\centering
\caption{Effect of retrieval method on \method.}
\label{tab:rag_comparison}
\small
\setlength{\tabcolsep}{5pt}
\renewcommand{\arraystretch}{1.1}
\begin{tabular}{lcc}
\toprule
\textbf{Retrieval} & \textbf{All Sections Proven} & \textbf{Sections Compiled} \\
\midrule
Hybrid & 123/300 (41.0\%) & 608/1393 (43.6\%) \\
BM25   & 126/300 (42.0\%) & 559/1393 (40.1\%) \\
Dense  & 134/300 (44.7\%) & 597/1393 (42.9\%) \\
\bottomrule
\end{tabular}
\end{table}

Table~\ref{tab:rag_comparison} compares retrieval methods using two complementary metrics. The first metric, \emph{All Sections Proven}, measures end-to-end mechanization success: a sketch is counted as successful only when all of its sections are formalized and compiled in dependency order. The second metric, \emph{Sections Compiled}, measures local section-level success across all generated sections.

Dense retrieval achieves the best end-to-end mechanization performance, proving all sections for $134$ out of $300$ sketches ($44.7\%$). BM25 and Hybrid also perform competitively, proving $126$ and $123$ sketches, respectively. At the section level, however, Hybrid compiles the largest number of individual sections, with $608$ compiled sections out of $1393$ ($43.6\%$), slightly higher than Dense with $597/1393$ ($42.9\%$). This shows that local section-level compilation does not always translate into full-sketch mechanization, since a sketch is counted as fully mechanized only if all of its sections compile together in dependency order.

\begin{table}[t]
\centering
\caption{Retrieval performance by target construct kind.}
\label{tab:rag_kind_breakdown}
\small
\setlength{\tabcolsep}{4pt}
\renewcommand{\arraystretch}{1.1}
\begin{tabular}{lccc}
\toprule
\textbf{Kind} & \textbf{Hybrid} & \textbf{BM25} & \textbf{Dense} \\
\midrule
Definition & 112/278 (40.3\%) & 114/278 (41.0\%) & 125/278 (45.0\%) \\
Lemma      & 8/19 (42.1\%)    & 9/19 (47.4\%)    & 8/19 (42.1\%) \\
Theorem    & 3/3 (100.0\%)    & 3/3 (100.0\%)    & 1/3 (33.3\%) \\
\bottomrule
\end{tabular}
\vspace{-4mm}
\end{table}

Table~\ref{tab:rag_kind_breakdown} further breaks down performance by target construct kind. Dense retrieval performs best on definition sketches, proving $125$ out of $278$ cases ($45.0\%$), suggesting that semantic retrieval is effective for matching informal descriptions to \prosa definitions. However, for proof-bearing constructs, the trend is different. Combining lemmas and theorems, BM25 proves $12$ out of $22$ proof-bearing sketches ($54.5\%$), Hybrid proves $11$ out of $22$ ($50.0\%$), and Dense proves $9$ out of $22$ ($40.9\%$). This suggests that sparse and hybrid retrieval are particularly useful when proof completion depends on exact lemma names, theorem identifiers, or library-specific proof patterns. Thus, while dense retrieval gives the strongest end-to-end performance overall, lexical retrieval remains important for proof-bearing constructs where exact \prosa references are often needed.

Overall, dense retrieval is most effective for semantic alignment, whereas sparse and hybrid retrieval remain valuable for proof-bearing constructs that depend on exact \prosa references.

\begin{rqanswerbox}
\noindent\textbf{Observation III.}
The choice of retrieval method affects both section-level compilation and end-to-end mechanization. Dense retrieval achieves the best overall result, fully mechanizing $134/300$ sketches ($44.7\%$), while Hybrid compiles the most individual sections, with $608/1393$ compiled sections ($43.6\%$). This suggests that semantic retrieval is especially useful for completing full dependency chains, while sparse retrieval remains useful for exact identifier and lemma-name matches.
\end{rqanswerbox}

%% file: sections/discussion.tex
\section{Conclusion}
\label{sec:conclusion}
This paper introduced \method, an LLM-assisted framework for generating mechanized \prosa/\rocq scripts for schedulability analyses in real-time systems literature. \method combines dependency-aware informal sketch extraction, retrieval from processed \prosa documentation, staged skeleton generation, and proof completion to guide LLMs toward \prosa-aware mechanization. 
Our evaluation shows that direct prompting of state-of-the-art LLMs is insufficient for reliable \prosa generation, while \method achieves a success rate of $44.7\%$ with dense retrieval. 
We also construct a mechanization-oriented corpus of informal sketches with dependency information. The corpus can facilitate future research on LLM-assisted mechanization of schedulability analyses, as domain-specific datasets for \prosa/\rocq-based schedulability analysis are currently lacking and remain a major bottleneck for automated script generation. 

As a research prototype, \method still faces challenges with deeper dependency chains and proof-bearing constructs that require precise \prosa context. Future work will improve retrieval, incorporate richer proof-state feedback, and extend the framework to broader classes of schedulability analyses. Moreover, we will study the capabilities of LLMs to generate schedulability constraints for new scheduling problems, for which \method can be used to mechanically verify the correctness of LLM-generated schedulability results.

%% file: sections/appendix.tex
\subsection{Illustrative Example with \textsc{Worst-Case Timing Requirements of Real-Time Tasks with Time Redundancy}}
 
\subsubsection{Extracted JSON and Informal Sketch for Illustrative Example}
\label{app:illustrative_example_informal_sketch}

\noindent Listing ~\ref{lst:json_example} presents the extracted json that is used for the Section ~\ref{sec:motivating_example}.

\begin{lstlisting}[ caption={Structured extraction of Definition~1 and Claim~1}, label={lst:json_example}]
[
  {
    "type": "definition",
    "identifier": "Definition 1",
    "formal_description": {
      "statement": "W_i_k_i = W_i_0 + A_i_k_i",
      "variables": {
        "W_i_k_i": "Worst case timing requirement of task
                     tau_i in the presence of k_i faults",
        "W_i_0": "Failure-free computational requirement
                   (WCET) of task tau_i",
        "A_i_k_i": "Worst case value of additional
                     reprocessing time and overhead to
                     tolerate k_i faults"
      },
      "assumptions": [
        "Faults are transient or intermittent",
        "Faults are detected immediately upon occurrence"
      ],
      "conclusion": "Defines the total time a task requires
                     to complete its execution and recovery
                     actions."
    },
    "informal_sketch": {
      "intuition": "The total time a task needs is its
                    normal execution time plus the maximum
                    possible time spent on fault detection,
                    recovery, and re-execution.",
      "steps": [
        "Identify the base execution time without faults
         W_i_0",
        "Calculate the maximum overhead A_i_k_i based on
         the specific redundancy technique used",
        "Sum them to find the total timing requirement"
      ],
      "key_insights": [
        "Separates the functional execution time from the
         fault-tolerance overhead"
      ]
    },
    "dependencies": []
  },
  {
    "type": "claim",
    "identifier": "Claim 1",
    "formal_description": {
      "statement": "W_i_k_i = k_i * (R_star + W_i_0)
                    + W_i_0",
      "variables": {
        "W_i_k_i": "Worst case timing requirement for
                     retry",
        "k_i": "Number of faults",
        "R_star": "Task restart overhead (constant)",
        "W_i_0": "Failure-free WCET"
      },
      "assumptions": [
        "Task must restart from the very beginning after
         every fault"
      ],
      "conclusion": "Calculates the WCTR for the Retry
                     redundancy technique."
    },
    "informal_sketch": {
      "intuition": "If a fault occurs, the task loses all
                    progress and must restart. In the worst
                    case, a fault occurs just before
                    completion, requiring a full re-execution
                    plus restart overhead for every fault.",
      "steps": [
        "For each of the k_i faults, add the cost of a
         full restart (R_star) and a full re-execution
         (W_i_0)",
        "Add the final successful execution time (W_i_0)"
      ],
      "key_insights": [
        "Retry is the most expensive technique because it
         discards all work done prior to the fault"
      ]
    },
    "dependencies": ["Definition 1"]
  }
]
\end{lstlisting}

\noindent Listing ~\ref{lst:informal_sketch} presents the informal sketch that is used for the Section ~\ref{sec:motivating_example}. This sketch is formatted as structured
\rocq comments and serves as the direct input to the skeleton code
generation phase.
 
\begin{lstlisting}[language=ML, caption={Informal sketch used as input to skeleton code generation}, label={lst:informal_sketch}]
(*
====section====
definition Definition 1
 
Statement:
W_i_k_i = W_i_0 + A_i_k_i
 
Variables:
W_i_k_i: Worst case timing requirement of task
          tau_i in the presence of k_i faults
W_i_0: Failure-free computational requirement
        (WCET) of task tau_i
A_i_k_i: Worst case value of additional
          reprocessing time and overhead to
          tolerate k_i faults
 
Assumptions:
1. Faults are transient or intermittent
2. Faults are detected immediately upon occurrence
 
Conclusion:
Defines the total time a task requires to complete
its execution and recovery actions.
 
Intuition for generating code:
The total time a task needs is its normal execution
time plus the maximum possible time spent on fault
detection, recovery, and re-execution.
 
Steps for generating code:
1. Identify the base execution time without faults
   W_i_0
2. Calculate the maximum overhead A_i_k_i based on
   the specific redundancy technique used
3. Sum them to find the total timing requirement
 
Key Insights:
1. Separates the functional execution time from the
   fault-tolerance overhead
*)
 
(*
====section====
claim Claim 1
 
Statement:
W_i_k_i = k_i * (R_star + W_i_0) + W_i_0
 
Variables:
W_i_k_i: Worst case timing requirement for retry
k_i: Number of faults
R_star: Task restart overhead (constant)
W_i_0: Failure-free WCET
 
Assumptions:
1. Task must restart from the very beginning after
   every fault
 
Conclusion:
Calculates the WCTR for the Retry redundancy
technique.
 
Intuition for generating code:
If a fault occurs, the task loses all progress and
must restart. In the worst case, a fault occurs
just before completion, requiring a full
re-execution plus restart overhead for every fault.
 
Steps for generating code:
1. For each of the k_i faults, add the cost of a
   full restart (R_star) and a full re-execution
   (W_i_0)
2. Add the final successful execution time (W_i_0)
 
Key Insights:
1. Retry is the most expensive technique because it
   discards all work done prior to the fault
*)
\end{lstlisting}

\subsection{IEEE Xplore API query}
\label{app:ieee_query}
We queried IEEE Xplore using schedulability-analysis and real-time-systems keywords.

\begin{center}
\small
\begin{minipage}{0.95\columnwidth}
\ttfamily
("schedulability analysis" OR "schedulability" OR "response time analysis" OR "response-time analysis") \\
AND ("real-time system" OR "real-time systems" OR "real-time task" OR "real-time scheduling" OR "worst-case execution time")
\end{minipage}
\end{center}

The API also allows us to restrict the search to specific conferences. For this purpose, we selected leading conferences in real-time systems and system automation. The conferences specified in our search are as follows:

\begin{center}
\small
\begin{minipage}{0.95\columnwidth}
\ttfamily
("Real-Time Systems Symposium", "RTSS"), 
("Real-Time and Embedded Technology and 
Applications Symposium", "RTAS"), 
("Euromicro Conference on Real-Time Systems", 
"ECRTS"), 
("Embedded Software", "EMSOFT"), 
("Real-Time Computing Systems and 
Applications", "RTCSA"), 
("Real-Time Networks and Systems", "RTNS"), 
("Robotics: Science and Systems", "RSS"), 
("Intelligent Robots and Systems", "IROS"),
\end{minipage}
\end{center}

\subsection{Prosa Documentation Details}
\begin{table}[H]
\centering
\caption{Role of major \textsc{Prosa} documentation modules.}
\label{tab:prosa_module_roles}
\small
\begin{tabularx}{\columnwidth}{lX}
\toprule
\textbf{Module} & \textbf{Role in Documentation Processing} \\
\midrule
Behavior & Core system semantics and basic concepts \\
Model & Task, processor, priority, and scheduler assumptions \\
Analysis & Reusable lemmas and intermediate proof developments \\
Results & Complete verified schedulability theorems and proof examples \\
Implementation & Executable instances and concrete schedulers \\
Util & Mathematical lemmas, helper functions, and tactics \\
\bottomrule
\end{tabularx}
\end{table}

\subsection{Pseudocode for skeleton code generation}
The pseudocode is presented in Algorithm~\ref{alg:skeleton_generation}

\begin{algorithm}[h]
\small
\caption{Phase 1: Skeleton Code Generation}
\label{alg:skeleton_generation}
\begin{algorithmic}[1]
\Require Sketch sections $\mathcal{S}$, retry budget $M_s$
\Ensure Skeleton script $P$

\State $P \gets \emptyset$ \Comment{accumulated script}

\For{each section $s_i \in \mathcal{S}$}
    \State $C_i \gets \textsc{RetrieveContext}(s_i)$
    \State $k_i \gets \textsc{InferKind}(s_i)$
    \State $success \gets false$

    \For{$a \gets 1$ to $M_s$}
        \State $\pi_i \gets \textsc{BuildPrompt}(s_i, C_i, P, k_i)$
        \State $b_i \gets \textsc{GenerateBlock}(\pi_i)$
        \State $b_i \gets \textsc{Normalize}(b_i)$

        \If{$\neg \textsc{CheckSkeletonRules}(b_i)$}
            \State \textbf{continue}
        \EndIf

        \If{$\neg \textsc{CheckAssumptions}(b_i, s_i)$}
            \State \textbf{continue}
        \EndIf

        \State $P' \gets \textsc{Concat}(P, b_i)$

        \If{$\textsc{Compile}(P')$}
            \State $P \gets P'$
            \State $\textsc{UpdateContext}(P, b_i)$
            \State $success \gets true$
            \State \textbf{break}
        \EndIf
    \EndFor

    \If{$success = false$}
        \State \Return $\textsc{Failure}(s_i)$
    \EndIf
\EndFor

\State \Return $P$
\end{algorithmic}
\end{algorithm}

\subsection{Pseudocode for Batch Completion with Iterative Repair}
The pseudocode is presented in Algorithm~\ref{alg:batch_completion}

\begin{algorithm}[t]
\small
\caption{Phase 2a: Batch Completion with Compiler-Guided Repair}
\label{alg:batch_completion}
\begin{algorithmic}[1]
\Require Skeleton sections $\mathcal{B}$, retry budget $M_b$
\Ensure Completed script $P$

\State $P \gets \emptyset$ \Comment{completed script}

\For{each section $B_i \in \mathcal{B}$}
    \If{$\neg \textsc{HasAdmitted}(B_i)$}
        \State $P \gets \textsc{Concat}(P, B_i)$
        \State \textbf{continue}
    \EndIf

    \State $C_i \gets \textsc{RetrieveContext}(B_i)$
    \State $\pi_i \gets \textsc{BuildCompletionPrompt}(B_i, C_i, P)$
    \State $p_i \gets \textsc{GenerateProof}(\pi_i)$
    \State $B_i' \gets \textsc{ReplaceAdmitted}(B_i, p_i)$
    \State $P' \gets \textsc{Concat}(P, B_i')$
    \State $success \gets \textsc{Compile}(P')$

    \For{$a \gets 1$ to $M_b$}
        \If{$success$}
            \State \textbf{break}
        \EndIf

        \State $e_i \gets \textsc{ExtractError}(P')$
        \State $\rho_i \gets \textsc{BuildRepairPrompt}(B_i', C_i, e_i)$
        \State $p_i \gets \textsc{RepairProof}(\rho_i)$
        \State $B_i' \gets \textsc{ApplyRepair}(B_i', p_i)$
        \State $P' \gets \textsc{Concat}(P, B_i')$
        \State $success \gets \textsc{Compile}(P')$
    \EndFor

    \If{$success$}
        \State $P \gets P'$
        \State $\textsc{MarkCompleted}(B_i')$
    \Else
        \State $P \gets \textsc{Concat}(P, B_i)$
        \State $\textsc{MarkFailed}(B_i)$
    \EndIf
\EndFor

\State \Return $P$
\end{algorithmic}
\end{algorithm}

\subsection{Illustrative Example with \emph{``Preemptively Scheduling Hard-Real-Time Sporadic Tasks on One Processor''}~\cite{illustrative_example}}
\label{app:illustrative_example_2}
 
We illustrate another workflow of \method using a concrete schedulability analysis drawn from the RTS literature, starting with the extraction of intermediate representations and leading to the final machine-verified \prosa script.

\subsubsection{Source Material and Extraction}

We consider the paper \emph{``Preemptively Scheduling Hard-Real-Time Sporadic Tasks on One Processor''}~\cite{illustrative_example} as a representative example. From this paper, we extract the following definitions and lemma. In particular, the lemma \texttt{Optimality of the Deadline Algorithm.} serves as the main target, as it establishes the optimality of the deadline algorithm for sporadic task systems. This lemma relies on several preceding modeling definitions.

\noindent \textbf{Definition~1 (Sporadic Task Model).}
A sporadic task is defined as
\[
  T_i = (e_i,d_i,p_i),
\]
where $e_i$ is the execution time, $d_i$ the relative deadline, and $p_i$ the
minimum separation between consecutive requests, with $e_i \le d_i$ and
$e_i \le p_i$. A task system is a finite set
$\tau = \{T_1,\ldots,T_n\}$.

\noindent \textbf{Definition~2 (Request Model).}
A request of task $T_i$ released at time $t_0$ is represented as $(i,t_0)$. It
requires $e_i$ units of processor time in the interval
\[
  [t_0,t_0+d_i),
\]
so its absolute deadline is $t_0+d_i$.

\noindent \textbf{Definition~3 (Legal Request Sets and Feasibility).}
A request set is legal if two requests of the same task are separated by at
least $p_i$:
\[
  |t_1-t_2| \geq p_i.
\]
A task system is feasible if every legal request set can be scheduled without
missing deadlines. Thus, feasibility is a universal property over all legal
sporadic arrivals.

\noindent \textbf{Definition~4 (Online Scheduling and Failure).}
An online scheduler decides at each time which active request executes. A
request $(i,t_0)$ is active at time $t$ if
\[
  t_0 \leq t < t_0+d_i
\]
and it has not yet received $e_i$ units of execution. The scheduler reports
failure when a request reaches its deadline without receiving enough execution.

\noindent \textbf{Definition~5 (Deadline Algorithm).}
The deadline algorithm selects, at each time, the active request with the
earliest absolute deadline. For two active requests $(i,t_1)$ and $(j,t_2)$, it
chooses $(i,t_1)$ if
\[
  t_1+d_i < t_2+d_j.
\]
Ties are resolved using a fixed task-index order.   

\noindent \textbf{Lemma~1 (Optimality of the Deadline Algorithm).}
The lemma states that the deadline algorithm is optimal for sporadic task
systems. For any request set, if some feasible schedule exists, then the
deadline algorithm also constructs one; otherwise, it reports failure.
Therefore,
\[
\tau \text{ is feasible}
\iff
\begin{aligned}[t]
&\text{the deadline algorithm succeeds} \\
&\text{for every legal request set}.
\end{aligned}
\]

\medskip
\noindent
These results are automatically extracted and structured using
LLM. Besides identifying the formal statement, variables,
assumptions, and conclusion of each invariant, the extraction also
produces an \emph{informal sketch}. The Informal Sketch ~\ref{lst:informal_sketch_illustrative_2} was extracted for this proof construct

\begin{lstlisting}[language=ML, caption={Extracted Informal Sketch for Optimality of the Deadline Algorithm.}, label={lst:informal_sketch_illustrative_2}]

(*
====section====
definition Definition 1

Statement:
A sporadic task task_i is a triple task_i = e_i, d_i, p_i where e_i, d_i, and p_i are positive integers, e_i <= d_i, and e_i <= p_i. A sporadic task system task_system is a finite set task_system = task_1, task_2, ..., task_n.

Variables:
task_i: sporadic task i
task_system: set of sporadic tasks
e_i: execution time of task_i
d_i: relative deadline of task_i
p_i: minimum separation between successive requests of task_i
n: number of tasks

Assumptions:
1. single processor
2. preemptive scheduling
3. discrete time model
4. e_i, d_i, p_i are positive integers
5. e_i <= d_i
6. e_i <= p_i

Conclusion:
Defines the sporadic task model and the task-system model.

Intuition for generating code:
Each sporadic task can release jobs at arbitrary times, but not too frequently. The parameter p_i prevents infinitely dense releases, while e_i and d_i describe how much processor time each released job needs and when it must complete.

Steps for generating code:
1. Represent each task task_i by execution time e_i, deadline d_i, and minimum separation p_i.
2. Require e_i <= d_i so that each individual job can fit inside its own deadline window.
3. Require e_i <= p_i so that the task does not request more execution than its minimum inter-arrival spacing can plausibly support.
4. Collect all tasks into task_system.

Key Insights:
1. The minimum separation p_i is the key distinction between sporadic and arbitrary aperiodic arrivals.
2. The model allows d_i > p_i, so jobs of the same task may have overlapping deadline windows.

*)

(*
====section====
definition Definition 2

Statement:
Let P = lcm(p_1, p_2, ..., p_n). A request of task_i at time t_0 is represented by request_i_t_0 = i, t_0. The request requires e_i units of processor allocation in interval [t_0, t_0 + d_i).

Variables:
P: least common multiple of all minimum separations p_i
request_i_t_0: request of task_i released at time t_0
t_0: release time of a request
e_i: execution requirement of task_i
d_i: relative deadline of task_i
p_i: minimum separation of task_i

Assumptions:
1. task_i belongs to task_system
2. t_0 >= 0
3. time is discrete

Conclusion:
Defines task requests and their execution windows.

Intuition for generating code:
A sporadic task generates individual requests or jobs. A request released at t_0 must receive e_i units of service before its absolute deadline t_0 + d_i.

Steps for generating code:
1. Take a task task_i and release time t_0.
2. Construct request_i_t_0.
3. Set the absolute deadline to t_0 + d_i.
4. Require the scheduler to allocate e_i time units inside [t_0, t_0 + d_i).

Key Insights:
1. Schedulability is checked over requests, not just over task parameters.
2. The absolute deadline is release_time plus relative deadline.

*)

(*
====section====
definition Definition 3

Statement:
A set of requests request_set is schedulable iff there exists a processor schedule that allocates e_i time units to every request request_i_t_0 in request_set within [t_0, t_0 + d_i). A set request_set is legal iff for any two requests request_i_t_1 and request_i_t_2 of the same task, abs(t_1 - t_2) >= p_i. The task_system is feasible iff every legal request_set is schedulable.

Variables:
request_set: set of task requests
request_i_t_0: request of task_i released at t_0
t_1: release time of one request
t_2: release time of another request
p_i: minimum separation of task_i
task_system: sporadic task system

Assumptions:
1. single processor
2. preemptive scheduling
3. all requests satisfy their task parameters

Conclusion:
Defines schedulability of a request set, legality of arrivals, and feasibility of a sporadic task system.

Intuition for generating code:
A request set is legal if it respects the sporadic separation constraints. A task system is feasible only if every possible legal arrival pattern can be scheduled without missing deadlines.

Steps for generating code:
1. Check all pairs of requests of the same task.
2. If any two releases are closer than p_i, the request set is illegal.
3. If the request set is legal, ask whether a valid preemptive single-processor schedule exists.
4. The task system is feasible only when every legal request set has such a schedule.

Key Insights:
1. Feasibility is a universal property over all legal arrival sequences.
2. The difficulty comes from the unbounded number of possible legal sporadic request sets.

*)

(*
====section====
definition Definition 4

Statement:
An online scheduling algorithm U maps each request_set and time t to either a selected active request request_i_t_0, an idle decision, and optionally failure. A request request_i_t_0 is active at time t iff t_0 <= t < t_0 + d_i and the request has not yet received e_i units of processor allocation in [t_0, t). U reports failure at time t iff there exists request_i_t_0 such that t_0 + d_i = t and the request has received less than e_i units in [t_0, t).

Variables:
U: online scheduling algorithm
request_set: set of requests presented to U
t: current time
request_i_t_0: request of task_i released at t_0
active: predicate indicating that a request is pending and before its deadline
failure: event indicating a missed deadline

Assumptions:
1. requests are presented to U at their release times
2. U is online and iterative
3. preemption is allowed at integer time boundaries

Conclusion:
Defines online scheduling, active requests, and failure.

Intuition for generating code:
At each time, the scheduler either runs one active request or idles. Failure occurs exactly when a request reaches its deadline without having received enough execution.

Steps for generating code:
1. At each time t, identify all active requests.
2. Choose one active request to execute or leave the processor idle.
3. Update the amount of service received by the chosen request.
4. If any request reaches its deadline without receiving e_i service, report failure.

Key Insights:
1. Failure is defined at the deadline boundary.
2. The active-request definition captures unfinished jobs that are still eligible to execute.

*)

(*
====section====
definition Definition 5

Statement:
The deadline algorithm U allocates the processor at time t to the active request with the nearest absolute deadline. For active requests request_i_t_1 and request_j_t_2, U chooses request_i_t_1 over request_j_t_2 if t_1 + d_i < t_2 + d_j, or if t_1 + d_i = t_2 + d_j and i < j.

Variables:
U: deadline algorithm
request_i_t_1: active request of task_i released at t_1
request_j_t_2: active request of task_j released at t_2
t_1_plus_d_i: absolute deadline of request_i_t_1
t_2_plus_d_j: absolute deadline of request_j_t_2

Assumptions:
1. preemptive scheduling
2. single processor
3. ties are broken by lower task index

Conclusion:
Defines the deadline-driven scheduling algorithm used throughout the paper.

Intuition for generating code:
The deadline algorithm is earliest-deadline-first with a deterministic tie-breaking rule. The request whose deadline is closest gets the processor.

Steps for generating code:
1. At time t, collect all active requests.
2. Compute each active request's absolute deadline.
3. Select the request with the smallest absolute deadline.
4. If two requests have the same absolute deadline, choose the one with smaller task index.

Key Insights:
1. The algorithm is EDF specialized to the paper's request model.
2. Tie-breaking does not affect whether failure occurs, but makes the schedule deterministic.

*)

(*
====section====
lemma Lemma 1

Statement:
The deadline_algorithm_U is optimal for sporadic task systems. Given any request_set, U constructs a schedule for request_set if one exists; otherwise U reports failure at some time. Therefore task_system is feasible iff U constructs a schedule for every legal request_set.

Variables:
deadline_algorithm_U: earliest-deadline scheduling algorithm
request_set: set of requests
task_system: sporadic task system

Assumptions:
1. single processor
2. preemptive scheduling
3. sporadic request model
4. legal request sets respect minimum separations

Conclusion:
EDF-style deadline scheduling is sufficient to decide feasibility over legal request sets.

Intuition for generating code:
For preemptive uniprocessor scheduling, always running the active job with the earliest deadline is optimal: if any schedule can meet all deadlines, the deadline algorithm can also meet them.

Steps for generating code:
1. Consider any legal request set.
2. Run the deadline algorithm on that request set.
3. If any feasible schedule exists, the deadline algorithm also succeeds.
4. If the deadline algorithm fails, no feasible schedule exists for that request set.
5. Thus task_system is feasible exactly when the deadline algorithm never fails on any legal request set.

Key Insights:
1. This lemma lets the paper focus on one canonical scheduler rather than all possible schedules.
2. It converts feasibility into absence of failure under the deadline algorithm.

*)

\end{lstlisting}

The extraction is also \emph{dependency-aware}. For each invariant, we
record its dependencies on previously introduced invariants, inducing a
partial order for formalization. An invariant is formalized only after
all dependencies have been processed. For example, Lemma~1 depends on
Definitions~1--5, which must therefore be formalized first. Since
\textsc{Prosa}/\rocq requires all referenced constructs to be defined
before use, dependency-aware ordering is necessary for successful
compilation.
 
\subsubsection{Retrieval-Augmented Generation}
 
Given the informal sketch of an invariant, the next step is to translate
it into a \textsc{Prosa} script. To support this translation, we employ
retrieval-augmented generation (RAG) over the \textsc{Prosa} codebase.
We use the \emph{statement}, \emph{conclusion}, and \emph{intuition}
fields from the informal sketch as a joint query and retrieve the
top-$k$ results from two complementary sources: (i)~\emph{example
scripts} that demonstrate similar constructs or proof patterns, and
(ii)~\emph{documentation fragments} describing the syntax, semantics,
and usage conventions of relevant \textsc{Prosa} modules. Together,
these provide the LLM with sufficient context to generate correct and
idiomatic code.
 
\subsubsection{Skeleton Code Generation}
 
Using the retrieved context, we prompt the LLM to generate
\emph{skeleton code}—a structurally complete, type-checkable
\textsc{Prosa} script in which all imports, section boundaries,
type-class contexts, variable declarations, and definition bodies are
fully specified, while every proof obligation is replaced by
\texttt{Admitted.} Listing~\ref{lst:skeleton_2} shows the skeleton code
generated for the motivating example.
 
\begin{lstlisting}[language=ML, basicstyle=\ttfamily\footnotesize,caption={Skeleton code generated from the informal sketch. All structural elements are fully elaborated; the proof body is deferred via \texttt{Admitted.}}, label={lst:skeleton_2}]

From mathcomp Require Import all_ssreflect.
Require Import prosa.model.task.concept.
Require Import prosa.model.task.arrival.sporadic.
Require Import prosa.model.priority.edf.
Require Import prosa.model.processor.ideal.
Require Import prosa.model.readiness.basic.
Require Import prosa.analysis.definitions.schedulability.
Require Import prosa.analysis.facts.behavior.completion.
Require Export prosa.results.edf.optimality.

Section SporadicEDFOptimality.

  Context {Task : TaskType} `{TaskCost Task} `{TaskDeadline Task} `{SporadicModel Task}.
  Context {Job  : JobType} `{JobTask Job Task} `{JobArrival Job} `{JobCost Job} `{JobDeadline Job}.
  Variable ts : seq Task.   (* the sporadic task system tau = {T_1,...,T_n} *)
  Hypothesis H_job_deadline_def :
    forall j, job_deadline j = job_arrival j + task_deadline (job_task j).

  Definition valid_sporadic_task (tsk : Task) : Prop :=
    [/\ 0 < task_cost tsk,
        0 < task_deadline tsk,
        0 < task_min_inter_arrival_time tsk,
        task_cost tsk <= task_deadline tsk &
        task_cost tsk <= task_min_inter_arrival_time tsk ].

  Definition request_absolute_deadline (j : Job) : instant :=
    job_arrival j + task_deadline (job_task j).   (* = job_deadline j, by H_job_deadline_def *)

  Definition feasible : Prop :=
    forall arr_seq : arrival_sequence Job,
      valid_arrival_sequence arr_seq ->
      (forall tsk, tsk \in ts -> respects_sporadic_task_model arr_seq tsk) ->
      exists sched : schedule (ideal.processor_state Job),
        valid_schedule sched arr_seq /\ all_deadlines_of_arrivals_met arr_seq sched.

  Definition online_failure_at
      (arr_seq : arrival_sequence Job)
      (sched : schedule (ideal.processor_state Job)) (t : instant) : Prop :=
    exists j, arrives_in arr_seq j /\ job_deadline j = t /\ ~~ completed_by sched j t.

  Definition deadline_algorithm_U (sched : schedule (ideal.processor_state Job)) : Prop :=
    EDF_schedule sched.

  Lemma deadline_algorithm_optimal :
    feasible <->
    (forall arr_seq : arrival_sequence Job,
       valid_arrival_sequence arr_seq ->
       (forall tsk, tsk \in ts -> respects_sporadic_task_model arr_seq tsk) ->
       exists sched : schedule (ideal.processor_state Job),
         valid_schedule sched arr_seq /\
         all_deadlines_of_arrivals_met arr_seq sched /\
         deadline_algorithm_U sched).
  Proof. Admitted.

End SporadicEDFOptimality.
\end{lstlisting}

The skeleton maps each extracted section to a corresponding \textsc{Prosa}
construct. Definition~1 is encoded as \texttt{valid\_sporadic\_task}, using
\texttt{task\_cost}, \texttt{task\_deadline}, and
\texttt{task\_min\_inter\_arrival\_time} for $e_i$, $d_i$, and $p_i$.
Definition~2 is encoded as \texttt{request\_absolute\_deadline}, representing
the absolute deadline $t_0+d_i$. Definition~3 is encoded as
\texttt{feasible}, which quantifies over all legal arrival sequences.
Definition~4 is encoded as \texttt{online\_failure\_at}, capturing missed
deadlines, and Definition~5 is encoded as \texttt{deadline\_algorithm\_U},
identified with \textsc{Prosa}'s \texttt{EDF\_schedule}. Lemma~1 is encoded as
\texttt{deadline\_algorithm\_optimal}, stating that feasibility is equivalent
to success of the deadline algorithm on every legal request set.

The surrounding type-class context, including \texttt{TaskType},
\texttt{JobType}, and \texttt{TaskCost}, connects these constructs to
\textsc{Prosa}'s scheduling framework. The \texttt{Admitted.} directive is used
only as a temporary placeholder, allowing the skeleton to type-check before the
proof is synthesized, with the final proof expected to rely on
\textsc{Prosa}'s \texttt{EDF\_optimality} theorem.
 
\subsubsection{Proof Completion}
In the final phase, we prompt the LLM to discharge the
\texttt{Admitted} obligations using the complete skeleton, retrieved RAG
context, and an instruction to complete a specific proof block. Providing
the full skeleton gives the model access to surrounding definitions,
instances, and auxiliary lemmas, improving proof generation. For this
example, the LLM produces the following proof in Listing ~\ref{lst:proof}:
 
\begin{lstlisting}[language=ML, basicstyle=\ttfamily\footnotesize, caption={Completed proof for the  Optimality of the Deadline Algorithm}, label={lst:proof}]
    Lemma deadline_algorithm_optimal :
    feasible <->
    (forall arr_seq : arrival_sequence Job,
       valid_arrival_sequence arr_seq ->
       (forall tsk, tsk \in ts -> respects_sporadic_task_model arr_seq tsk) ->
       exists sched : schedule (ideal.processor_state Job),
         valid_schedule sched arr_seq /\
         all_deadlines_of_arrivals_met arr_seq sched /\
         deadline_algorithm_U sched).
  Proof.
    rewrite /feasible /deadline_algorithm_U.
    split.
    - move=> FEAS arr_seq VALID SPORADIC.
      apply: EDF_optimality.
      have [sched [VALID_SCHED DEADLINES_MET]] :=
        FEAS arr_seq VALID SPORADIC.
      by exists sched.
    - move=> EDF_FEAS arr_seq VALID SPORADIC.
      have [sched [VALID_SCHED [DEADLINES_MET EDF_SCHED]]] :=
        EDF_FEAS arr_seq VALID SPORADIC.
      by exists sched.
  Qed.
\end{lstlisting}

The proof follows the two directions of the equivalence. First,
\texttt{rewrite /feasible /deadline\_algorithm\_U} unfolds the extracted
definitions, reducing the statement to the relationship between general
feasibility and existence of an EDF schedule. In the forward direction, the
assumption \texttt{FEAS} provides some valid schedule that meets all deadlines
for any legal arrival sequence. The proof then applies \texttt{EDF\_optimality},
which states that if such a feasible schedule exists, then an EDF schedule also
exists.

In the reverse direction, the assumption already provides a valid schedule that
meets all deadlines and additionally satisfies \texttt{EDF\_schedule}. Since
feasibility only requires the existence of some valid schedule meeting all
deadlines, the EDF-specific condition is discarded. Thus, the completed proof
shows that the extracted lemma \texttt{deadline\_algorithm\_optimal} is a valid
\textsc{Prosa} formalization of the paper's claim that the deadline algorithm
is optimal for preemptive uniprocessor sporadic task systems.
\medskip
\noindent
This example demonstrates how our pipeline systematically transforms
informal real-time scheduling results into machine-verified
\textsc{Prosa} proofs through a structured sequence of extraction,
retrieval-augmented skeleton generation, and targeted proof completion.

\subsection{Distribution of system invariants and their number of dependencies in the generated System Invariant Dataset}
\label{app:invariant_dist}
\begin{figure}[t]
\centering
\includegraphics[width=\columnwidth]{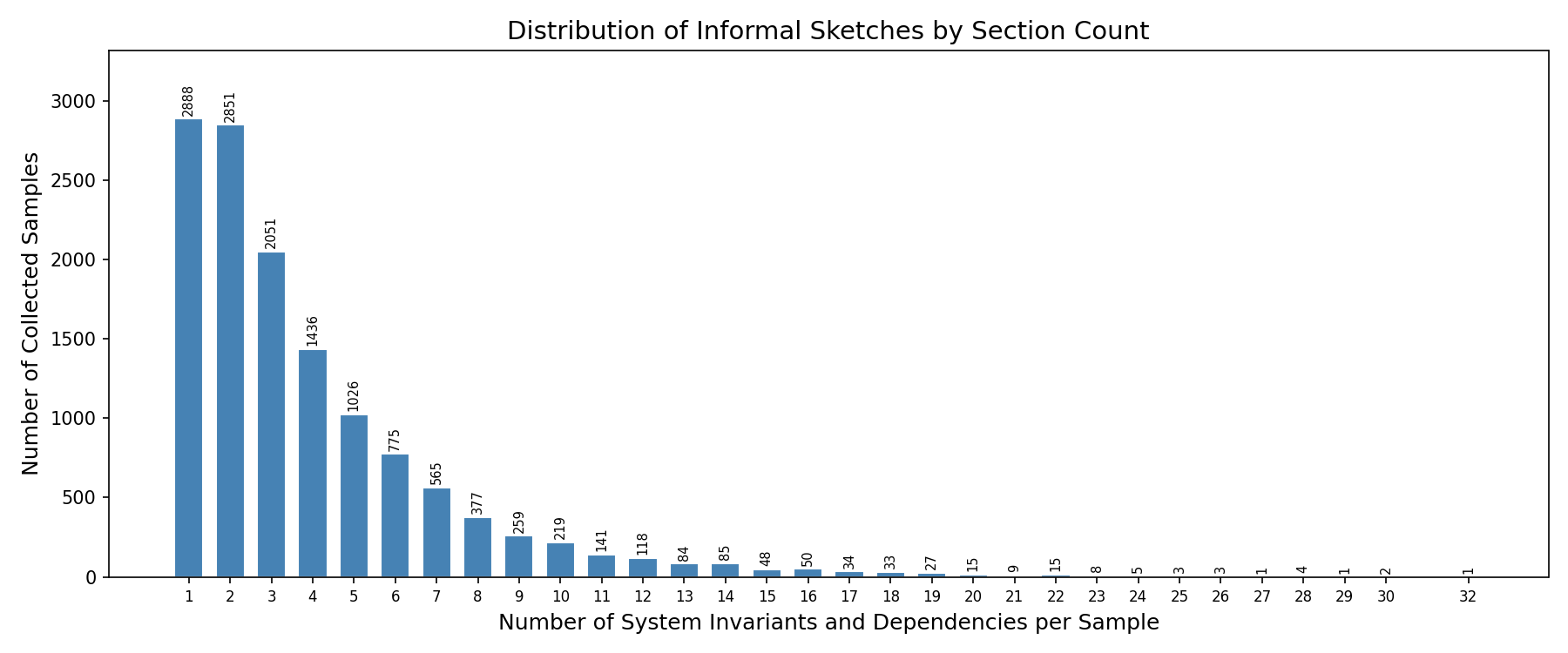}
\caption{Distribution of system invariants and their dependencies.}
\label{fig:invariant_distribution}
\end{figure}

Figure~\ref{fig:invariant_distribution} shows the distribution of the total number of invariants per sketch, defined as the target invariant plus its associated dependencies. The frequency decreases as the number of dependencies increases, indicating that highly dependent invariants are less common in the extracted corpus.

\subsection{Fine-Grained Mapping from Raw Sketch Kinds to \rocq Keywords}
\label{app:raw_sketch_mapping}

Table~\ref{tab:raw_sketch_mapping} presents the complete mapping from raw sketch kinds to their corresponding \rocq keywords, along with the rationale used in our normalization process.

\setlength{\LTleft}{0pt}
\setlength{\LTright}{0pt}

\small
\begin{longtable}{p{0.26\textwidth}p{0.16\textwidth}p{0.52\textwidth}}
\caption{Fine-grained mapping from raw sketch kinds to \rocq keywords.}
\label{tab:raw_sketch_mapping}\\
\toprule
\textbf{Raw Sketch Kind} & \textbf{\rocq Keyword} & \textbf{Reason} \\
\midrule
\endfirsthead

\caption[]{Fine-grained mapping from raw sketch kinds to \rocq keywords (continued).}\\
\toprule
\textbf{Raw Sketch Kind} & \textbf{\rocq Keyword} & \textbf{Reason} \\
\midrule
\endhead

\bottomrule
\endfoot

\bottomrule
\endlastfoot

\multicolumn{3}{l}{\textbf{Definition family}} \\
definition & Definition & Direct equivalent \\
formula & Definition & A named mathematical formula = constant definition \\
equation & Definition & A named equation = defined constant/relation \\
implicit\_definition & Definition & Implicitly defined concept \\
derived\_definition & Definition & Derived from other definitions \\
foundational\_equation & Definition & Named base equation = constant \\
rta\_equation & Definition & Response-time analysis equation = named bound constant \\
calculation\_formula & Definition & A formula used in calculation = named constant \\
derived\_formula & Definition & Derived formula = defined constant \\
resource\_planning\_formula & Definition & Formula for planning = named constant \\
objective\_function & Definition & Optimization objective = a function definition \\
optimization\_objective & Definition & Same as objective\_function \\
derived\_function & Definition & Derived non-recursively (not Fixpoint) \\
function & Definition & Non-recursive function \\
formulation & Definition & Mathematical formulation = a definition \\
task\_set\_definition & Definition & Explicit definition of a task set \\
informal\_definition & Definition & Informal but definitional in nature \\

\midrule
\multicolumn{3}{l}{\textbf{Fixpoint family}} \\
algorithm & Fixpoint & Algorithms are recursive = Fixpoint \\
algorithmic\_sketch & Fixpoint & Sketch of a recursive algorithm \\
informal\_algorithmic\_sketch & Fixpoint & Informal version of algorithmic\_sketch \\
algorithm\_sketch & Fixpoint & Same as algorithmic\_sketch \\
algorithm\_component & Fixpoint & Part of a recursive algorithm \\
algorithm\_definition & Fixpoint & Defines a recursive algorithm \\
algorithm\_description & Fixpoint & Describes an algorithm \\
algorithmic\_derivation & Fixpoint & Algorithmically derived result \\
algorithmic\_definition & Fixpoint & Formally defined algorithm \\
transformation\_algorithm & Fixpoint & An algorithm that transforms its input \\
method & Fixpoint & Computational method = procedure \\
procedure & Fixpoint & Explicit computational procedure \\
heuristic & Fixpoint & Iterative/recursive computational procedure \\
recurrence & Fixpoint & Recurrence relation = recursive definition \\

\midrule
\multicolumn{3}{l}{\textbf{Lemma family}} \\
lemma & Lemma & Direct equivalent \\
claim & Lemma & A claim is an auxiliary provable fact \\
proposition & Lemma & Proposition = Lemma in \rocq convention \\
observation & Lemma & An observation is an informal lemma \\
property & Lemma & A provable proposition \\
statement & Lemma & A mathematical statement to be proved \\
formal\_statement & Lemma & Formal version of a statement \\
condition & Lemma & A provable condition/predicate \\
constraint & Lemma & Proved inequality or bound \\
calculation & Lemma & Result of a calculation = proved equality \\
schedulability\_test & Lemma & Decidable test = proved as lemma \\
derived\_test & Lemma & A test derived from other results \\
optimization\_constraint & Lemma & Proved inequality/bound \\
optimization\_claim & Lemma & Claim about an optimisation \\
application & Lemma & Instantiation of a result \\
invariant & Lemma & System/loop invariants proved as lemmas \\
remark & Lemma & Remarks are informal lemmas \\
fact & Lemma & A mathematical fact \\
result & Lemma & A proved result \\
restriction & Lemma & A restriction = conditional lemma \\
conjecture & Lemma & Unproved; placeholder with Admitted \\
property/constraint & Lemma & Combined property/constraint \\
problem & Lemma & Decision/optimisation problem \\
rule & Lemma & Inference rule = proved as lemma \\
optimization\_problem & Lemma & Proved bound/property \\
schedulability\_condition & Lemma & Schedulability condition = proved predicate \\
problem\_statement & Lemma & Problem stated formally \\
inequality & Lemma & A proved inequality \\
known\_result & Lemma & Existing proved result \\
principle & Lemma & e.g. scheduling principle = proved property \\
derived\_rta & Lemma & Response-time bounds are proved, not defined \\
extension & Lemma & Extends an existing lemma \\
transformation & Lemma & Proved equivalence under transformation \\
refinement & Lemma & Proved relationship between specifications \\
informal\_sketch & Lemma & Informal sketch of a proof obligation \\
informal & Lemma & Generic informal mathematical statement \\
policy & Lemma & Scheduling policy property = provable proposition \\
incomplete\_malformed\_statement & Lemma & Safe fallback \\

\midrule
\multicolumn{3}{l}{\textbf{Theorem / Corollary / Hypothesis families}} \\
theorem & Theorem & Direct equivalent \\
corollary & Corollary & Direct equivalent \\
hypothesis & Hypothesis & Direct equivalent \\
assumption & Hypothesis & Explicit assumption in a \rocq Section context \\

\end{longtable}

\twocolumn